\documentclass[runningheads]{llncs}
\usepackage{underscore}
\usepackage{graphicx}
\usepackage{amsmath,amssymb} 
\usepackage{color}
\usepackage{algorithm}
\usepackage{algpseudocode}
\usepackage{epsfig} 
\usepackage{epstopdf} 
\usepackage{wrapfig}
\usepackage{kantlipsum}
\usepackage[width=122mm,left=12mm,paperwidth=146mm,height=193mm,top=12mm,paperheight=217mm]{geometry}

\begin{document}
\newcommand*\samethanks[1][\value{footnote}]{\footnotemark[#1]}
\pagestyle{headings}
\mainmatter

\title{Spatial Attention Deep Net with Partial PSO
for Hierarchical Hybrid Hand Pose Estimation} 

\titlerunning{Spatial Attention Deep Net with Partial PSO for Hierarchical Hybrid Hand}

\authorrunning{Qi Ye, Shanxin Yuan, Tae-Kyun Kim}

\author{Qi Ye\thanks{indicates equal contribution}, Shanxin Yuan\samethanks, Tae-Kyun Kim}


\institute{ Department of Electrical and Electronic Engineering,\\
	 Imperial College London\\
	\email{ \{q.ye14,s.yuan14,tk.kim\}@imperial.ac.uk}
}

\maketitle

\begin{abstract}
Discriminative methods often generate hand poses kinematically implausible, then generative methods are used to correct (or verify) these results in a hybrid method. Estimating 3D hand pose in a hierarchy, where the high-dimensional output space is decomposed into smaller ones, has been shown effective. Existing hierarchical methods mainly focus on the decomposition of the output space while the input space remains almost the same along the hierarchy. In this paper, a hybrid hand pose estimation method is proposed by applying the kinematic hierarchy strategy to the input space (as well as the output space) of the discriminative method by a spatial attention mechanism and to the optimization of the generative method by hierarchical Particle Swarm Optimization (PSO). The spatial attention mechanism integrates cascaded and hierarchical regression into a CNN framework by transforming both the input(and feature space) and the output space, which greatly reduces the viewpoint and articulation variations. Between the levels in the hierarchy, the hierarchical PSO forces the kinematic constraints to the results of the CNNs. The experimental results show that our method significantly outperforms four state-of-the-art methods and three baselines on three public benchmarks.

\end{abstract}

\section{Introduction}
\begin{figure*}
\centering  

\includegraphics[width= 1.00\textwidth,page=1,trim=1cm 0cm 0cm 0cm,clip=true]{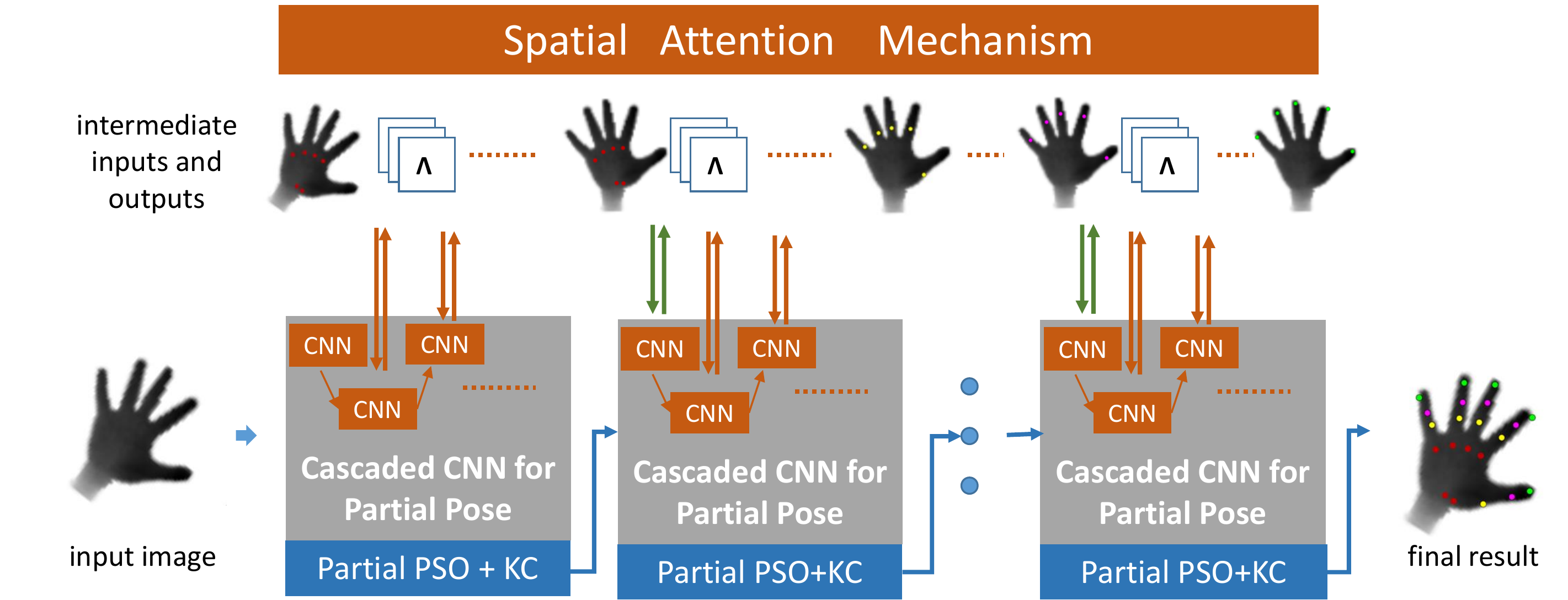}	
  \caption{Structure of the proposed method. The Spatial Attention Mechanism integrates
  the cascaded and hierarchical hand pose estimation into one framework. The hand pose is estimated layer by layer in the order of the articulation complexity, with the spatial attention module to transform the input$/$feature and output space. Within each layer, the partial pose is iteratively refined both in viewpoint and location with the spatial attention module, which leads both the feature and output space to a canonical one. After the refinement, the partial PSO is applied to select estimations within the hand kinematic constraints (short as KC in the figure) among the results of the cascaded estimation. $\Lambda$ denotes the CNN feature maps.
  } 
  \label{fig:overview}
\end{figure*}

The problem of 3D hand pose estimation can be formulated as the configuration of the variables representing a hand model given depth images. The problem is challenging with complicated variations caused by high Degree of Freedom (DoF) articulations, multiple viewpoints, self-similar parts, severe self-occlusions, different shapes and sizes. With these variations, configurations of the hand variables given a depth image lie in a high-dimensional space.


Many prior works have achieved good performance by different methods~\cite{oikonomidis2011efficient,qian2014realtime,sharp2015accurate,oberweger2015training,neverova2015hand,tang2013real,keskin2012hand,Ionescu2014iter,liang2014parsing,rogez20143d,stenger2006model,ballan2012motion,intel,supancic2015depth,taylor2014user,krejov2015combining}. Among the discriminative methods that learn the mapping from the depth images to the hand pose configurations, Sun~{\it et al.}~\cite{sun2015cascade} refine the hand pose by two levels of a hierarchy (palm, and fingers) in a cascaded manner by viewpoint-invariant pixel difference features in random forest. 
Oberweger~{\it et al.}~\cite{oberweger2015hands} apply the cascaded method to CNN for iteratively refining partial poses, initialized by the full hand pose estimation. 

The discriminative and generative methods are combined in a hierarchy in the Hierarchical Sampling Optimization (HSO)~\cite{tang2015opening}. In each layer, random forests are first used to regress partial poses and a partial joint energy function is introduced to evaluate the results and select the best one to the next layer.
The hierarchical optimization with refinement that estimates the hand pose in the order of articulation complexity of the hand is a promising framework as the searching space is decomposed into smaller parts and the refinement leads to more accurate results.

However, the method in~\cite{sun2015cascade}  and the discriminative part of HSO~\cite{tang2015opening} only focus on breaking down the complexity in the output space hierarchically, i.e., decomposing the hand variables; in other words, the hierarchical strategy is carried out in the output space while the input space or the feature space stays the same along the hierarchy. For the cascaded refinement~\cite{sun2015cascade,oberweger2015hands}, the input or feature space is only partially updated with results from previous stages, either by cropping or rotating, and the features~\cite{sun2015cascade} are computed on the original whole images in each iteration. In addition, the optimization of the energy function is performed in a brute force way in~\cite{tang2015opening}.

In this paper, we propose a hybrid method with iterative (cascade) refinement for hand pose estimation, illustrated in Fig 1, which not only applies the hierarchical strategy to the output space but also the feature space of the discriminative part and the optimization of the energy function of the generative part. 

For the discriminative part, a spatial attention mechanism is introduced to integrate cascaded (with multiple stages) and hierarchical (with multiple layers) regression into a CNN framework by transforming both the input (and feature space) and the output space. In the transformed space, the viewpoint and articulation variations of the feature space and the output space is largely reduced, which greatly simplifies the estimation. Along the hierarchy, with the spatial attention mechanism, the features for the initial stage of each layer are spatially transformed from input images based on the estimation results of the last stage of the previous layer. Within each layer, the features are iteratively updated by the spatial attention mechanism. By this dynamic spatial attention mechanism, not only the most relevant features for the hand variable estimation are selected but also the features are transformed to a canonical, expected viewpoint gradually, which simplifies the estimations in the following stages and layers. As such, discriminative features are extracted for each partial pose estimation in each iteration. In this way, we learn a deep net with spatial transformation tailored towards our hand pose estimation problem.

In the generative part, the optimization organized in the hierarchy prevents error accumulation from previous layers. Between the levels of the hierarchy, partial PSO with a new energy function is incorporated to enforce hand kinematic constraints. It generates samples under the Gaussian distribution centered on the results of the discriminative method, and selects estimations within the hand kinematic constraints. The search space of the generative method is largely reduced by estimating partial poses. 

To evaluate our method, extensive experiments have been conducted on three public benchmarks. The experimental results show that our method significantly outperforms state-of-the-art methods on these datasets.

\section{Related Work}
\label{sec:rw}

\subsubsection{Feature Selection with Attention}
\label{ssec:sa}

Learning or selecting transformation-invariant representations or features by neural networks has been studied in many prior works and among them, attention mechanism has gained much attention in object recognition and localization recently. Girshick {\it et al.}~\cite{girshick2014rich} produce region proposals as representations for CNN to focus its localization capacity on these regions instead of a whole image. DRAW~\cite{gregor2015draw} integrates a spatial attention mechanism mimicking that of human eye into a generative model to generate image samples in different transformations. Sermanet{\it et al.}~\cite{sermanet2014attention} use an attention model to direct a high resolution input to the most discriminative regions to do fine-grained categorization. An end-to-end spatial transformation neural network is proposed in~\cite{jaderberg2015spatial}.

The attention mechanism is tailored to our highly articulated problem by breaking down to the viewpoint and articulation complexity in a hierarchy and refining estimation results in a cascade. The hierarchical structure with cascade refinement enables us to use a spatial transformation to not only select most relevant features as in prior works aforementioned and also transform the feature and the output space into a new one which leads to our expected, canonical space.

\subsubsection{Cascaded and Hierarchical Estimation}
\label{ssec:cashier}
The cascaded regression strategy has shown good performances in the face analyses~\cite{zhao2014unified,zhu2015face}, human body estimation~\cite{dollar2010cascaded,toshev2014deeppose} and hand pose estimation~\cite{sun2015cascade,oberweger2015hands} and in most of these works, the features are hand-crafted, such as pixel difference features~\cite{dollar2010cascaded,sun2015cascade}, landmark distance features~\cite{zhao2014unified}, SIFT~\cite{xiong2013supervised}. Oberweger {\it et al.}~\cite{oberweger2015hands} use CNN to learn features automatically but with only partial spatial transformations by cropping input images and in another work~\cite{oberweger2015training}, they use the images generated by CNN as the feedback to refine the estimation. Sun {\it et al.}~\cite{sun2015cascade} refine the hand pose using pixel difference features updated for viewpoints of whole images in each iteration. The features in both works are partially transformed either by cropping patches from the input images or rotating features calculated from the whole image.
On the other hand, a hierarchical strategy that estimates hand poses in the order of hand articulation complexity achieves good performance~\cite{tang2015opening,sun2015cascade}. HSO~\cite{tang2015opening} estimates partial poses separately in the kinematic hierarchy while the input space remains unchanged. Sun {\it et al.}~\cite{sun2015cascade} estimate partial poses holistically in two layers of a hierarchy by calculating rotation invariant pixel difference features from the whole image.

Our proposed method fully transforms the feature space and the output space together in both cascaded and hierarchical manner. For each iteration of the cascade, no new features are learned as features are obtained by a spatial transformation applied to the feature maps of an initial stage. For the hierarchy, only a small region which has been transformed to a canonical view is fed into CNN. In this way, the hierarchical and cascaded strategy is not only applied to the output space as in prior work but also the transformed input and feature space.

\subsubsection{Hybrid Methods}
\label{ssec:Hybrid}
A standard way of combining the discriminative methods and the generative methods is first providing candidate results by the discriminative methods, then using them as the initial state of the generative methods to optimize full hand poses~\cite{sharp2015accurate,sridhar2015fast,tompson2014real,krejov2015combining}, and it has demonstrated good performances. 
As discussed in the above, searching the full hand pose space has a high complexity. 
We adopt a partial pose optimization to reduce the complexity of each estimation, which is integrated into our hierarchical structure. HSO~\cite{tang2015opening} also has partial pose evaluations between the levels of a hierarchy but the evaluations are carried out in a brute-force way, while we propose a new kinematic energy function which is optimized by the partial PSO.

\section{Method overview}
\label{sec:overview}

\begin{wrapfigure}{R}{0.4\textwidth}
\centering    
\includegraphics[trim=4cm 8cm 10cm 2cm, clip=true,width=0.80\textwidth]{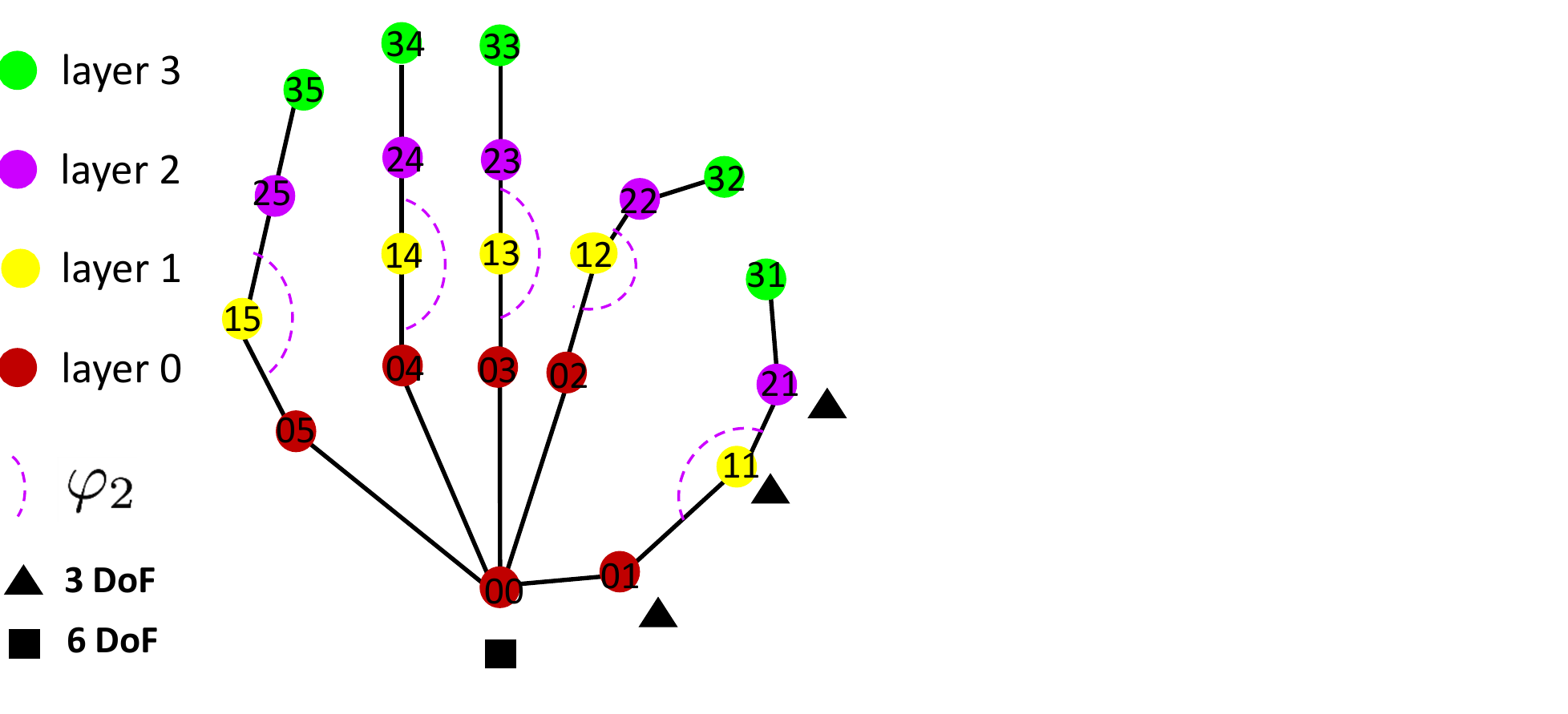}
\caption{Hand model. 21 joints are divided into four layers, each joint overlaid with index number. ${\varphi}_2$ is the bone rotations for five joints in Layer 2.}
\label{fig:handmodel}
\end{wrapfigure}

Hand pose estimation is to estimate the 3-D locations of the hand's 21 key joints $S$ given depth image $I$, which is normalized by the size of the depth image. The ground truth of $S$ is denoted as ${S}^{\ast}$. In our approach we divide the 21 joints into four layers ${\{{S}_{l}\}}_{l=0}^{3}$ where the value of $l$ is also the order of our hierarchical estimation, see Fig~\ref{fig:handmodel}. For each layer $l$, $j$ is used to denote a single joint on one finger, in the order from thumb to pinky with the number starting from 1 to 5 (for the wrist joint in the first layer, $j$ is 0). With all the definitions, the hand variables to be estimated are expressed as ${\{{\{{S}_{lj}\}}_{j=0}^{5}\}}_{l=0}\cup{\{{\{{S}_{lj}\}}_{j=1}^{5}\}}_{l=1}^{3}$.
 
Our method estimates (and trains) 4 layers sequentially with the spatial attention mechanism(see Sec~\ref{ssec:st}) linking the layers by transforming the input (and feature) and output space interactively and partial PSO enforcing kinematic constraints to the CNN prediction, which is shown in Fig~\ref{fig:overview}. In each layer $l$, the estimation is refined iteratively by learning the residual of the ground truth ${S}_{lj}^{\ast}$ to the results ${S}_{lj}^{k-1}$ of the previous stage, where $k$ denotes the ${k}^{th}$cascaded stage(for details, see Sec~\ref{ssec:cad}). The spatial transformation modules are applied to the feature maps from the initial stage of the cascade and the outputs of stage $k-1$ to get aligned attention features and output space for the learning of residual of stage $k$.

The result ${S}_{lj}^{{K}_{l}}$ of the final stage ${K}_{l}$ is fed into the post-optimization process using PSO for initialization. The partial PSO(see Sec~\ref{sec:ppso}) is introduced to enforce kinematic constraints to the results from the cascaded estimation and refine the partial pose. Along with PSO, we adopt the hand bone model (Fig~\ref{fig:handmodel}), which has 51 DoFs: layer 0 has 6 DoFs, denoting the global orientation (represented by a 4-D unit quaternion) and global location (3 DoFs); each of layer 1, 2, 3, has 15 DoFs, denoting the five bone rotations. Our hand model fixes the six joints on the palm and keeps the bone lengths of the fingers.

The optimal of the PSO is passed to layer $l+1$. Before the estimation of the next layer, the spatial attention mechanism is applied on input images and estimation results of current layer (and the ground truth for next layer during training).

\section{Partial Pose Estimation by Spatial Attention Deep Net}
\label{sec:cashier}

\subsection{Spatial Attention Mechanism for Hand Space}
\label{ssec:st}
\begin{figure*}
\centering
	\includegraphics[width=12cm,page=1,trim=0cm 0.5cm 0.5cm 0.5cm,clip=true]{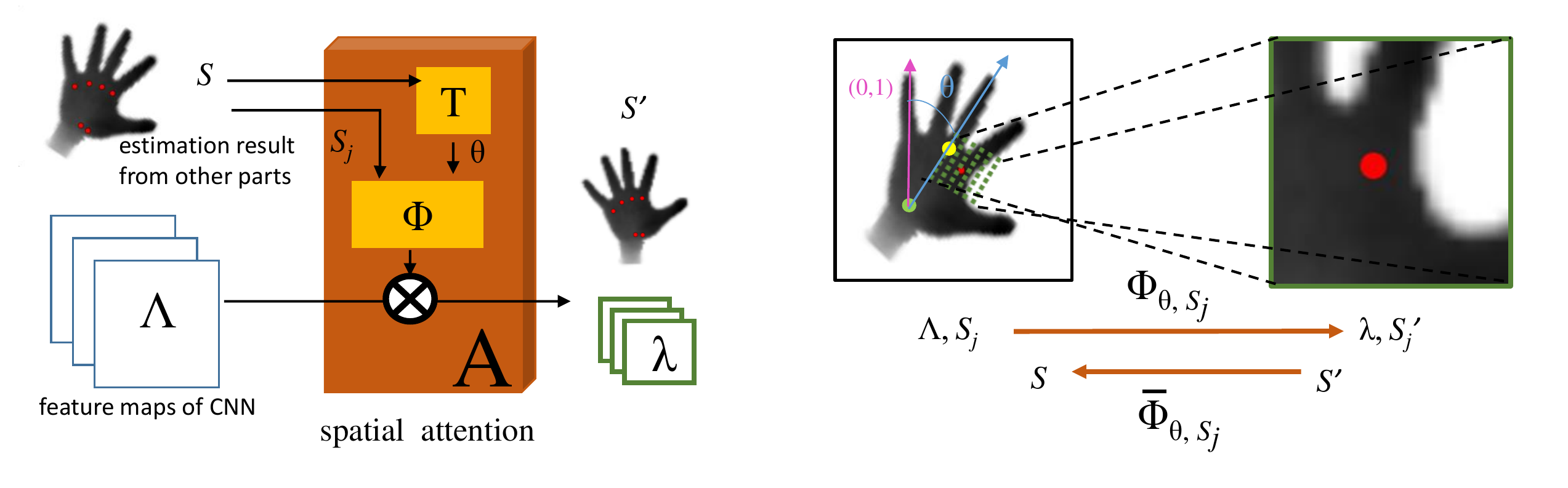}
  \caption{Spatial attention mechanism. Left: the spatial attention module is split into the calculation of rotation $T$ and the spatial transformation $\Phi$. Right: the mapping between input feature maps and output features maps. For clarity, we use hand images to represent the feature maps. Both the feature maps, estimation results (and ground truth in training) are transformed to a new space by ${\Phi}_{\theta,{S}_{j}}$. The locations can be transformed back by the inverse function $\bar{{\Phi}}_{\theta,{S}_{j}}$.} 
  \label{fig:st}
\end{figure*}

Before the elaboration of the hand pose estimation, the mechanism of spatial attention is explained. For notational simplicity, we skip the layer index $l$ and the stage index $k$ as the mechanism is applied to all layers and all stages similarly. The inputs of the spatial attention module are the estimation result of ${S}_{j}$ , where $j$ denotes ${j}_{th}$ joint in the layer, and the features maps of CNN (and input images), denoted by $\Lambda\in{\rm I\!R}^{ W\times H}$.

The spatial module $A$, illustrated in the left figure of Fig~\ref{fig:st}, can be split into two parts: the calculation of rotation $T$ and the pixel mapping $\Phi$. The global in-plane rotation $\theta$ (see the right figure of Fig~\ref{fig:st}) is the angle between the vector of the wrist joint (joint 00 in Fig.~\ref{fig:handmodel}) to the root joint of middle finger (joint 03 in Fig.~\ref{fig:handmodel}) in Layer 0 and the vector $(0,1)$ representing the upright hand pose and can be expressed as $\theta = T({S}_{3},{S}_{0})$. For the other layers $l$ ($l>0$), the rotation is obtained from Layer 0.

For the pixel mapping, displayed in the right figure of Fig~\ref{fig:st}, in which {\it pixel} here means an element of the feature maps (and input images), we use $({x}^{i},{y}^{i})$ to denote a pixel on the input feature map $\Lambda$  and $({x}^{o},{y}^{o})$ on the output feature maps  $\lambda\in{\rm I\!R}^{W'\times H'}$. For the deep features for joint $j$, the translation parameter is the xy coordinates of ${S}_{j}$ on the feature map($\Lambda$) coordinate system, i.e. $({t}_{x},{t}_{y})$. The mapping between $({x}^{i},{y}^{i})$ and $({x}^{o},{y}^{o})$ is 
\begin{align}
\begin{bmatrix}
    {x}^{i}  \\
    {y}^{i} \\
    1 
\end{bmatrix}
=
\begin{bmatrix}
    b\cdot\cos(\theta) & \quad b\cdot\sin(\theta) &  \quad  {t}_{x}  \\
    -b\cdot\sin(\theta) & \quad b\cdot\cos(\theta) & \quad   {t}_{y}  \\
\end{bmatrix}
\begin{bmatrix}
    {x}^{o}  \\
    {y}^{o} \\
    1 
\end{bmatrix}
\label{eq:trans}
\end{align}
\begin{align}
(\lambda,S') = {\Phi}_{\theta,{S}_{j}}(\Lambda,S) 
\label{eq:phi}
\end{align}
where $({x}^{i},{y}^{i})$ and $({x}^{o},{y}^{o})$ are normalized by its corresponding width and height of the input and output feature maps. $b$ is the ratio of the width of $\lambda$ to the width of $\Lambda$ (or the height as we keep the aspect ratio). If $b$ is 1, the transformation is rotation and translation. When $b$ is less than 1, the transformation allows cropping and the cropping size is the same as the size of the output feature maps $\lambda$.

Once we get the transformation parameters, the mapping between $\lambda$ and  $\Lambda$ are established by interpolating the pixel values. We also apply the transformation to the estimation results $S$ (and the ground truth ${S}^{\ast}$ in training) by Eq.\ref{eq:trans} , only on their $xy$ coordinates, the value of the $z$ coordinate remains unchanged. All the inputs are in a new coordinate system, or a new space. We use ${\Phi}_{\theta,{S}_{j}}$ in Eq.~\ref{eq:phi} to wrap the mapping function in Eq.~\ref{eq:trans} for all the pixels on the feature maps $\Lambda$ and also symbolize the transformation for $S$. $\bar{{\Phi}}_{\theta,{S}_{j}}$, denoting the inverse function of ${\Phi}_{\theta,{S}_{j}}$, acquired by replacing $\theta$ by $-\theta$, $({t}_{x},{t}_{y})$ by $-({t}_{x},{t}_{y})$ and $b$ by $1/b$ in Eq.~\ref{eq:trans}, transforms the output space of CNNs to the original one.

\subsection{Cascaded Regression within Each Hierarchical  Layer}
\label{ssec:cad}

\begin{figure*}
\centering
	\includegraphics[width=12cm,page=1,trim=1cm 0.9cm 0cm 0.5cm,clip=true]{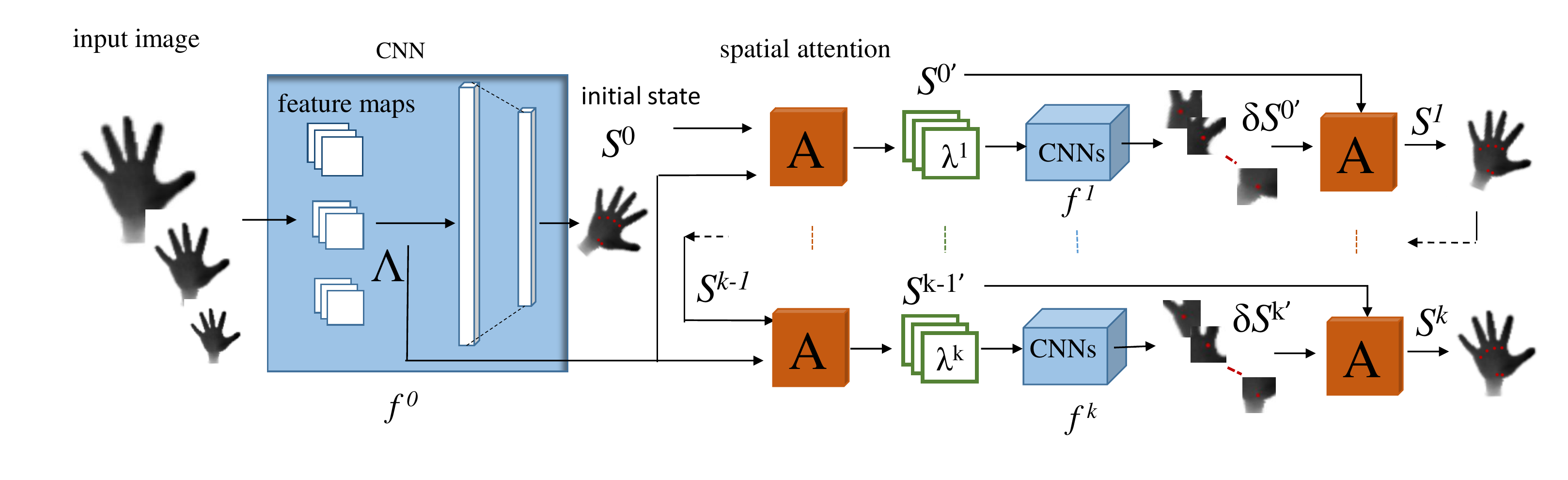}
  \caption{Cascaded partial pose estimation with spatial attention modules for Layer 0. The feature maps $\Lambda$ from the initial stage is transformed by spatial attention modules $A$ with estimation result ${S}^{k-1}$ form previous stage before feeding into the current stage $k$.} 
  \label{fig:casatt}
\end{figure*}

Within each layer of the hierarchy, the joint locations $\{{S}_{j}\}$ are estimated in a cascaded manner, shown in Fig.~\ref{fig:casatt}. We leave out the layer subscript $l$ as the cascaded regression is applied to all layers. At first, an initial CNN model ($\{{f}_{j}^{0}\}$) regresses the joint location $\{{S}_{j}\}$. It not only provides an initial state $\{{S}_{j}^{0}\}$ for the following iterative refinements but also deep feature maps $\Lambda$ for other regressors. In the following stages, the joint locations are refined iteratively. Between the refinement stages, the spatial attention modules $A$ transform the deep feature maps $\Lambda$ to a new space based on the estimation result $\{{S}_{j}^{k-1}\}$ from the previous stage to achieve viewpoint-invariant and discriminative features for the following regressors.  
 
For a certain joint $j$ in stage $k$, the features ${\lambda}_{j}^{k}$ is mapped from $\Lambda$ by ${\Phi}_{{\theta}^{k-1},{S}_{j}^{k-1}}$, where the ${S}_{j}^{k-1}$ is the result of the previous stage and ${\theta}^{k-1}$ is calculated by $T({S}_{0}^{k-1},{S}_{3}^{k-1})$ (the updating of $\theta$ happens only in Layer 0, and for other layers the value of $\theta$ is fixed after Layer 0). At the same time, the estimation result ${S}_{j}^{k-1}$ and the ground truth ${S}_{j}^{\ast}$ are both transformed by the module, resulting ${S}_{j}^{k-1'}$ and ${S}_{j}^{\ast'}$. Therefore, all the inputs for the regressor ${f}_{j}^{k}$ in stage $k$ that estimates the residual ${S}_{j}^{\ast'}- {S}_{j}^{k-1'}$ of joint $j$ are in a new space. After training or testing, the output of the regressor is then transformed back by $\bar{{\Phi}}_{\theta,{S}_{j}}$. For the joint $j$, the process is repeated until a satisfactory result is achieved (seen Sec.~\ref{sec:exp} for the choice of cascaded stages)and we use ${K}_{l}$ to denote the final stage for Layer $l$. For other joints, the refinement is carried out in parallel with the same process.

The above refinements for a single joint can be mathematically expressed as 

\begin{equation}
{(\lambda}_{j}^{k},{S}_{j}^{k-1'})={\Phi}_{{\theta}^{k-1},{S}_{j}^{k-1}}(\Lambda,{S}_{j}^{k-1})
\label{eq:sa}
\end{equation}
\begin{equation}
{S}_{j}^{k} = \bar{{\Phi}}_{{\theta}^{k-1},{S}_{j}^{k-1}}({f}_{j}^{k}({\lambda}_{j}^{k})+{S}_{j}^{k-1'})
\label{eq:update}
\end{equation}
where Eq.~\ref{eq:sa} is the spatial attention mechanism which transforms all the inputs of stage $k$ to a new space and Eq.~\ref{eq:update} estimates the residual ${\delta{S}}_{j}^{k'}$ by $ {f}_{j}^{k}({\lambda}_{j}^{k})$, updates the estimation by adding the residual estimated ${\delta{S}}_{j}^{k'}$ to the result from the previous stage ${S}_{j}^{k-1'}$, and transforms the added result back to the original space.

\subsection{Hierarchical Regression}
\label{ssec:hier}

For the regression in layer 0, all the joints in the initial stage are learned together in order to keep the kinematic constraints among them as the values of these joints are highly correlated. The input of the initializor ${f}_{0}^{0}$ is multi-resolution images $I$, the original image and the images downsampled from the original one by the factor of 2 and 4, the output is the joint locations. The input and feature space of the regressors for different joints in the cascaded stages are updated separately by the spatial function ${\Phi}_{{\theta}^{k-1},{S}_{0j}^{k-1}}$. The output of the regressor in the stage $k$ refines the estimation result ${S}_{0j}^{k-1}$ in the previous stage $k-1$ in a new space and are transformed back by $\bar{{\Phi}}_{{\theta}^{k-1},{S}_{0j}^{k-1}}$. The cascaded regression stop in stage ${K}_{0}$. The whole refinement stages are the same as in Se~\ref{ssec:cad}.

For the hierarchical estimation in layer $l (l>0)$, the inputs are multi-resolution input images $I$, the estimation result $\{{S}_{l-1,j}^{{K}_{l-1}}\}$ from the previous layer $l-1$ and the viewpoint estimation ${\theta}^{{K}_{0}}$ from layer 0. For notational simplicity, we denote ${\theta}^{{K}_{0}}$ as $\theta$ and skip the joint index $j$. $\theta$ is fixed for all layers ($l>0$) and the same process is applied to all the joints separately.

The input space for the initializor ${f}_{l}^{0}$ of layer $l$ is transformed from multi-resolution images $I$ by the spatial attention module. The mapping is 
\begin{align}
(I', {S}_{l-1}^{{K}_{l-1}'}, {S}_{l}^{\ast'})={\Phi}_{\theta,{S}_{l-1}^{{K}_{l-1}}}(I,{S}_{l-1}^{{K}_{l-1}}, {S}_{l}^{\ast})
\end{align}
so the input for the initializor ${{f}_{l}^{0}}$ is patches $I'$ cropped from multi-resolution input images ${I}$ centred at ${S}_{l-1}^{{K}_{l-1}}$ and its corresponding coordinates in the downsampled images, and rotated by $\theta$. The offset labels for training ${{f}_{l}^{0}}$ is $ {\Delta S}_{l}^{\ast'} = {S}_{l}^{\ast'} - {S}_{l-1}^{{K}_{l-1}'}$, which is equivalent to the sum of the ground truth offset ${S}_{l}^{\ast'}-{S}_{l-1}^{\ast'}$ and the remaining residual of the previous layer ${S}_{l-1}^{\ast'}-{S}_{l-1}^{{K}_{l-1}'}$. This implies the initializor ${{f}_{l}^{0}}$ not only predicts the joint offsets of the current layer to the  previous layer but also corrects the residual errors of the previous layer.

The initializor ${{f}_{l}^{0}}$ provides the initial offset state ${\Delta S}_{l}^{0'}$ and feature maps ${\Lambda}$ for the refinement stages. For the refinement stages, the procedure is the same as discussed in Sec~\ref{ssec:cad}. The only difference from Sec~\ref{ssec:cad} is that the viewpoint is static, whose value is the result of the final cascaded stage in Layer 0, and feature maps ${\Lambda}$ has already been transformed by rotation in the initial stage; thus for the stage $k$, the feature space is transformed and updated by the function ${\Phi}_{{S}_{l}^{k-1}}$(no rotation transformation) and the output space is transformed with ${\Phi}_{\theta,{S}_{l}^{k-1}}$ (rotation and translation transformation).

The parameter $b$ in the spatial attention module needs to be set. For the initial stage(except the initial stage of layer 0), it is set according to the offset range. All the ground truth ${S}_{l}^{\ast}$ and the estimation result ${S}_{l-1}^{{K}_{l-1}}$ of layer $l-1$ is first transformed by ${\Phi}_{\theta,{S}_{l-1}^{{K}_{l-1}}}$ with $b=1$ to get means along the $xy$ coordinates of the absolute value of the offsets for all the training data in the new space. $b$ is set to be two times of the larger offset mean divided by original image width $W$. For the refinement stages, they are set according to the residual range of the estimation results in the initial stage. All the ground truth and the estimation results of the initial stage are first transformed by ${\Phi}_{{\theta}^{0},{S}_{j}^{0}}$ (for layer $l (l>0)$, ${\theta}^{0}$ is the final estimation result ${\theta}^{{K}_{0}}$ of layer 0) with $b=1$ to get means along the $xy$ coordinates of the absolute value of the residuals for all the training data in the new space. As the feature maps are filtered by kernels, max-pooled, and have different resolutions but the ground truth are normalized according to original image size, ${S}_{j}^{0}$ and the mean of residual should also be changed with the kernel size, the pool size and the downsampling ratio to set the value of $b$.

\section{Partial PSO with Kinematic Constraints for Final Refinement}
\label{sec:ppso}

For each layer, based on our discriminative part's prediction, we do final refinement by explicitly introducing partial kinematic constraints with Particle Swarm Optimization. Particle Swarm Optimization (PSO) is a stochastic optimization algorithm introduce by Kennedy and Eberhart~\cite{488968} in 1995, originated in the social behaviors' studies of synchronous bird flocking and fish schooling. The original PSO algorithm has been modified by several researchers to improve its convergence properties and search capabilities. We adopt the variant of PSO with an inertia weight parameter~\cite{shi1998modified}.

\begin{figure}[t]
	\centering
		\includegraphics[trim=1cm 8.7cm 0.5cm 1.5cm, clip=true, width=1.00\textwidth]{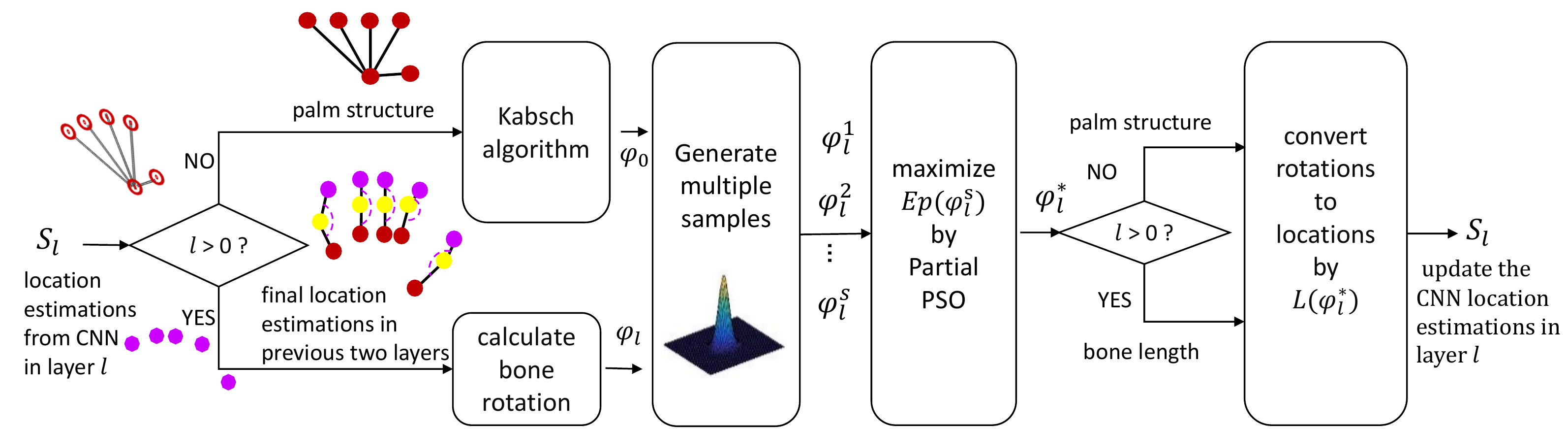}
	\caption{Pose refinement with partial PSO enforcing kinematic constraint. Given palm spatial structure and layer 0's location estimation by CNN, we first inference the $\varphi_{0}$ using Kabsch algorithm \cite{kabsch1976solution}, and then find $\varphi_{0}^{*}$ maximizing the energy function by partial PSO. The rotation  $\varphi_{0}^{*}$ is converted to locations using the palm structure to update the CNN estimation result.  For other partial pose $\varphi_{l}(l>0)$, the optimization is the same with layer 0 while the inference for the initialization of the optimization is calculating the bone rotation and the conversion back to locations uses the bone length. }
	\label{fig:PartialPOSEPSO}
\end{figure}

Our whole hand pose for PSO is defined as $\{\varphi_{0}, \varphi_{1},  \varphi_{2}, \varphi_{3}\}$, where $\varphi_{0} \in {\rm I\!R}^{7}$ and $\varphi_{l} (l = 1, 2, 3) \in {\rm I\!R}^{15}$ are our partial poses. $\varphi_{0} = \{q, x, y, z\}$, where $q$ is a 4-D unit quaternion~\cite{oikonomidis2011efficient,sharp2015accurate} representing the global rotation, $[x, y, z]$ is the global location of the whole hand. $\varphi_{l}$ denotes five 3D Euler angles in layer $l$, each angle representing a bone rotation which is the angle between the bone connecting the joint in layer $l$ and the corresponding joint in layer $l-1$, and the other bone connecting the joint in layer $l-1$ and the corresponding joint in layer $l-2$ (when $l-2<0$, the corresponding joint in layer $l-2$ is wrist). Fig.~\ref{fig:handmodel} demonstrates $\varphi_{2}$, five angles in layer $l=2$.

\subsubsection{Energy Function}
\label{ssec:Energy Function}
For each layer, PSO is used to estimate the final partial pose base on the inferred partial pose. We designed a new energy function that applied to partial pose and explicitly taking into account the kinematic constraints. Our energy function $Ep$ for each layer is as follows:
\begin{equation}
Ep(\varphi_{l}^{s}) = P(\varphi_{l}^{s})Q(\varphi_{l}^{s}),\quad\quad  
\end{equation}
where the first item, $P(\varphi_{l}^{s}) \propto N(\varphi_{l}^{s}; \varphi_{l}, \Sigma)$, is the prior probability of the $s_{th}$ Gaussian sample from mean $\varphi_{l}$, $\Sigma$ is a diagonal covariance matrix that is manually set to ensure that each parameter varies in valid ranges. $s = 1, 2, ..., N$ is the index of samples for each layer, we set $N = 100$ in our experiments. $P(\varphi_{0}^{s})$ encodes the spatial structure of the six joints on the palm and $P(\varphi_{l}^{s}) (l = 1, 2, 3)$ keeps the bone length information. 

To acquire the prior probability $P(\varphi_{0}^{s})$, we first choose Kabsch algorithm~\cite{kabsch1976solution} to find the optimal affine transformation matrix (global translation and rotation, i.e. $\varphi_{0}$ )from our hand model for the six joints on the palm to the CNN results, as shown in the top pipeline of ~Fig.~\ref{fig:PartialPOSEPSO}. The hand model for the palm joints can be seen as the palm joint locations of an upfront reference hand pose with wrist located on the original coordinate. By generating samples from Gaussian distribution centred on $\varphi_{0}$ instead of $S_{0}$ from CNN which usually violates the kinematic constraint, we get the $P(\varphi_{0}^{s})$ that keeps the spatial structure of palm joints.

For $P(\varphi_{l}^{s})$ of other layers $l>0$, we first get five bone rotations $\varphi_{l}$ by calculating the angles which are demonstrated in the bottom pipeline of Fig.~\ref{fig:PartialPOSEPSO} with joint estimation locations of CNN in current layer $l$ and the joint estimation locations from layer $l-1$ and $l-2$, and then sample from the Gaussian distribution centred on $\varphi_{l}$. When converting rotations to locations for evaluations of the second term, we enforce the constraint of the bone length.

The second item, $Q(\varphi_{l}^{s}) \propto \sum_{S_{lj}^{s} \in L(\varphi_{l})}[B(S_{lj}^{s}) + D(S_{lj}^{s})] $,  denotes the likelihood of all joint $\{S_{lj}^{s}\}$ belongs to the hand, where $L(\varphi_{l}^{s})$ converts rotations $\varphi_{l}^{s}$ into locations $\{S_{lj}^s\}$. Similar to Tang {\it et al.} \cite{tang2015opening} silver function, the term $B(S_{lj}^{s})$ forces each joint joint $S_{lj}^{s}$ to lie inside the hand silhouette. The term $D(S_{lj}^{s})$ makes sure joint $S_{lj}^{s}$ lies inside the depth range of a major point cloud.

\section{Experiment}
\label{sec:exp}

The evaluation of our proposed method is conducted on three publicly datasets. \textbf{ICVL}~\cite{tang2015opening} dataset is a real sequence captured by Intel RealSense with the range of view about 120 degrees consisting 1596 test frames and 16008 training frames. 16 bone centre locations are provided for each hand pose.
\textbf{NYU}~\cite{tompson2014real} dataset is a real sequence acquired by PrimeSense containing 8252 test-set and 72757 training-set frames with a full range of views.  36 joint locations are provided for each hand pose.
\textbf{MSRC}~\cite{sharp2015accurate} dataset is a challenging dataset that covers a full range of views and complex articulations with 100000 synthetic images in the training-set and 2000 synthetic images in the test set.  22 joint locations are provided for each hand pose. As the annotations of these datasets do not conform to each other, we use the annotation version in~\cite{tang2015opening} that labels locations of the joints as demonstrated in Fig~\ref{fig:handmodel}.


\begin{figure*}
	\begin{minipage}[b]{.24\linewidth}
	{\includegraphics[width=3.cm,trim=0.5cm 0cm 2cm 0.5cm,clip=true]{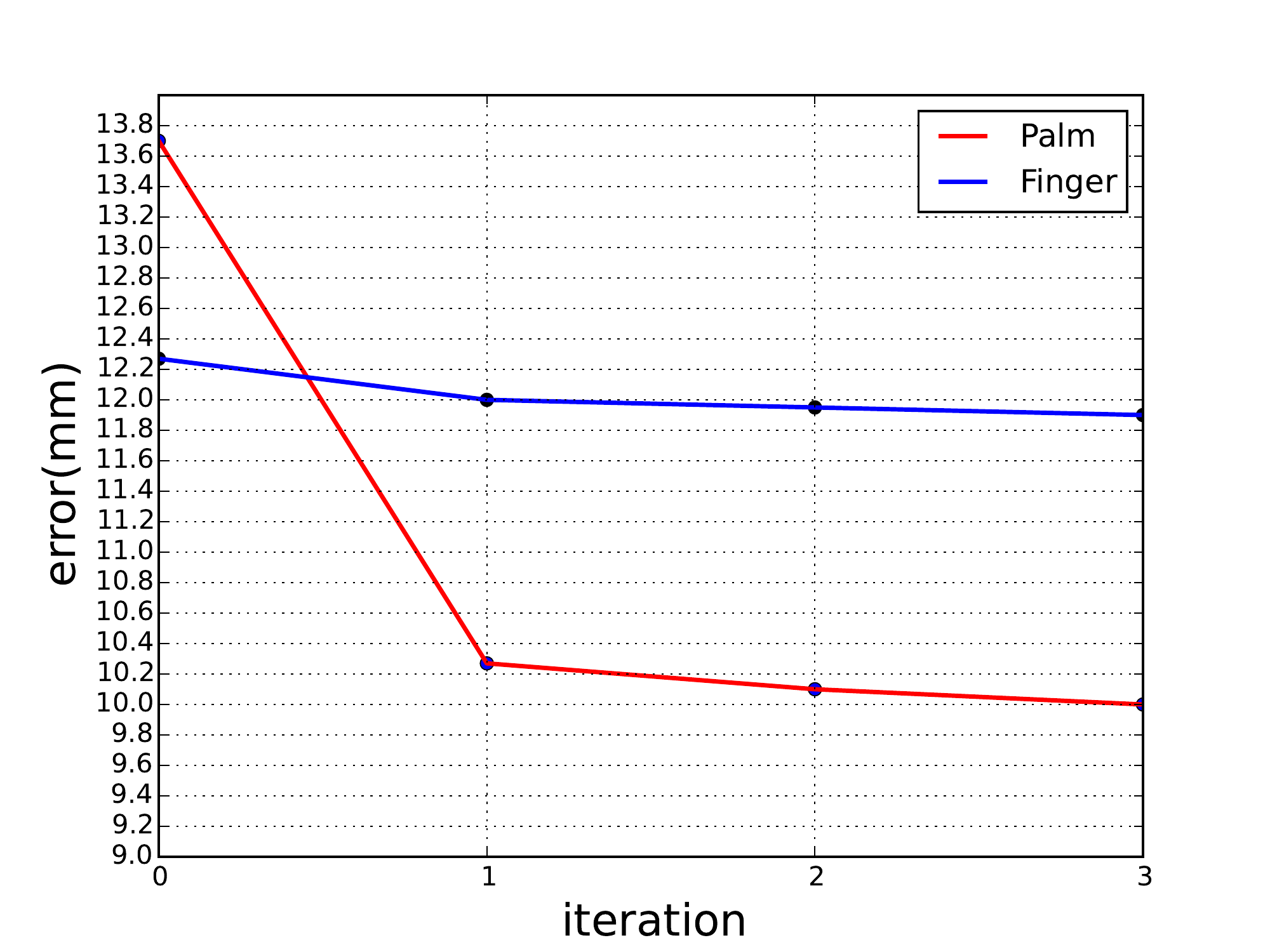}}
	\end{minipage}
	\begin{minipage}[b]{.24\linewidth}
	{\includegraphics[width=3.cm,trim=0.5cm 0cm 2cm 0.5cm,clip=true]{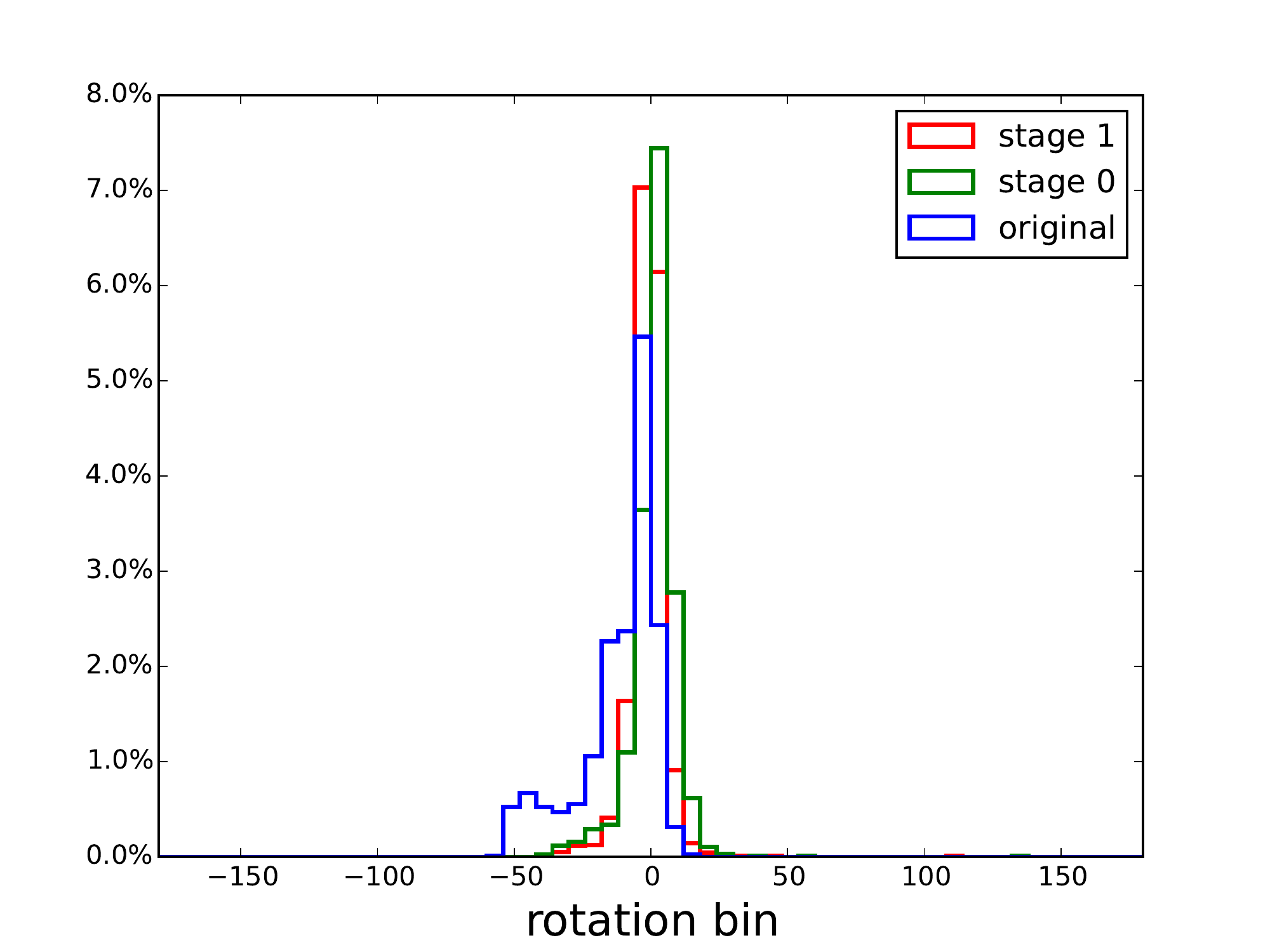}}
	\end{minipage}
	\begin{minipage}[b]{.24\linewidth}
	{\includegraphics[width=3.cm,trim=0.5cm 0cm 2cm 0.5cm,clip=true]{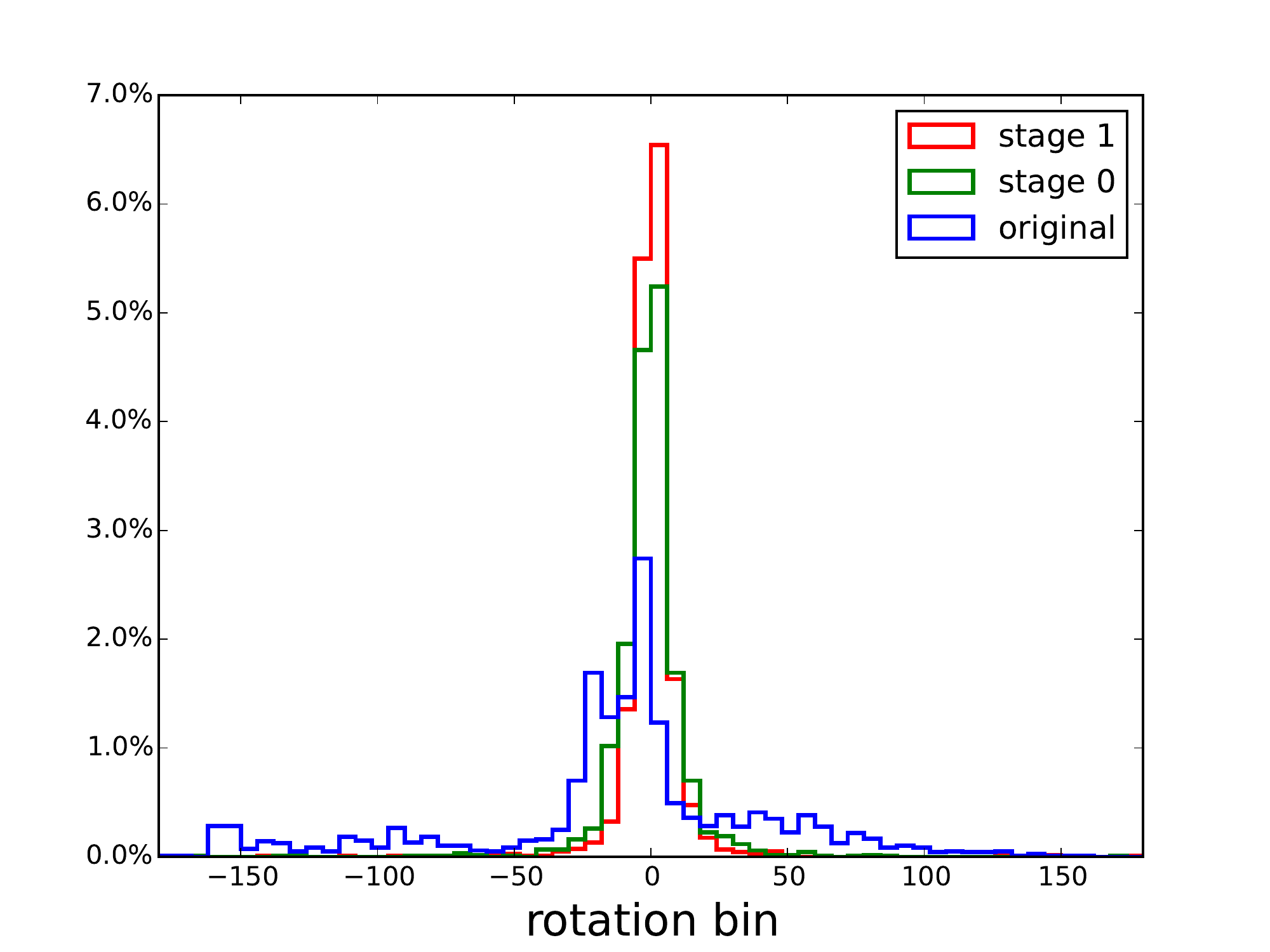}}
	\end{minipage}
	\begin{minipage}[b]{.24\linewidth}
	{\includegraphics[width=3.cm,trim=0.5cm 0cm 2cm 0.5cm,clip=true]{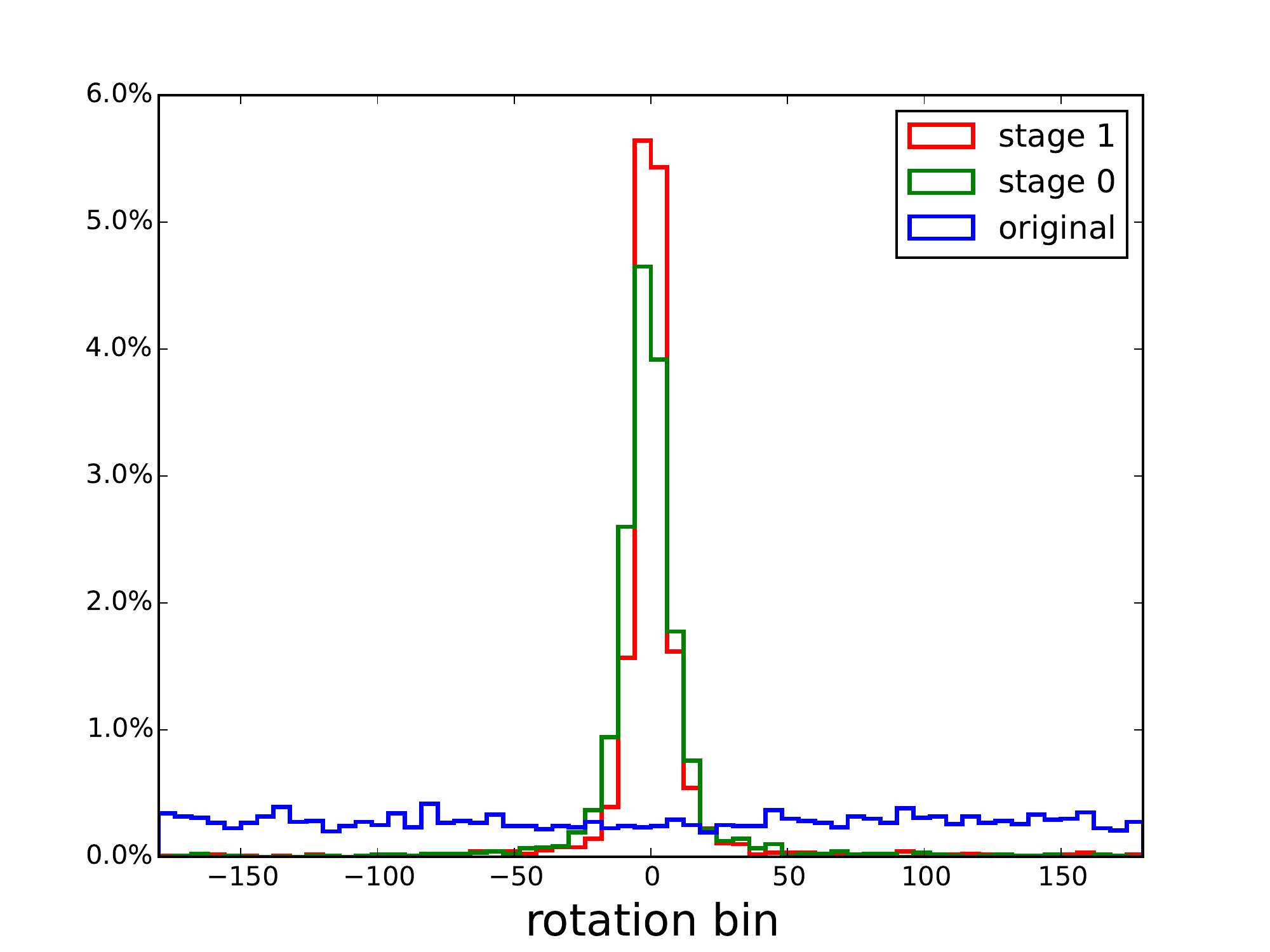}}
	\end{minipage}		

\caption{ First: Errors for a joint on the palm ${S}_{00}$ and a joint on the middle finger ${S}_{13}$ for 4 stages. Second, third and forth: In-plane viewpoint distribution of testing set for different stages on ICVL, NYU, MSRC respectively. The blue, green and red line corresponds to the in-plane rotation distribution of original ground truth, ground truth rotated after initial stage and the first stage. The rotation estimation error after the initial stage and the first stage is 5.9 and 4.4 for ICVL, 8.0 and 6.1 for NYU, 10.9 and 9.2 for MSRC in the unit of degree.}
\label{fig:iteration}
\end{figure*}

We compare the results of different methods by the proportion of joints within a certain maximum error of the distance of the predicted results to the ground truth~\cite{sharp2015accurate}.  We set the number of iterations $K$ on the observation of the error saturates after a certain stage in the cascaded stages, shown in the first figure of Fig. ~\ref{fig:iteration}. We set ${K}_{0}$ for Layer 0 to 1 and ${K}_{l},l>0$ to 0, which gives us a good balance between the accuracy and the memory consumption. All the experiments are run with Intel i7, 24GB RAM and NVIDIA GeForce GTX 750 Ti. The structures for our CNN models are implemented by Theano~\cite{2016arXiv160502688short} and the details are provided in the supplementary material. For our partial PSO, we generates $100$ samples for each layer, and iterate $5$ generations.

\subsection{Self-comparison}
\label{ssec:selfcomp}

To evaluate our proposed method (Hybr_Hier_SA) and the discriminative part Hier_SA, we implement three baselines. The first baseline (Holi) estimates the whole hand pose with a single CNN. The second baseline(Holi_Derot) consists of two steps: one step predicting the in-plane rotation of the hand pose by a CNN and rotating the hand pose to upright view; the other step estimating the whole hand pose by another CNN. The third one (Holi_SA) is a holistic cascaded regression network without hierarchy, which initializes the whole hand pose with a CNN and refines the hand pose joint by joint via spatial attention mechanism by a set of CNNs. For fair comparison, we set the size of the parameters of the methods to be roughly the same: the parameters are stored in 32 bit float and the size of parameters is 130MB.
\begin{figure*}
	\begin{minipage}[b]{.325\linewidth}
	{\includegraphics[width=4.cm,trim=0.5cm 0cm 1.8cm 1.3cm,clip=true]{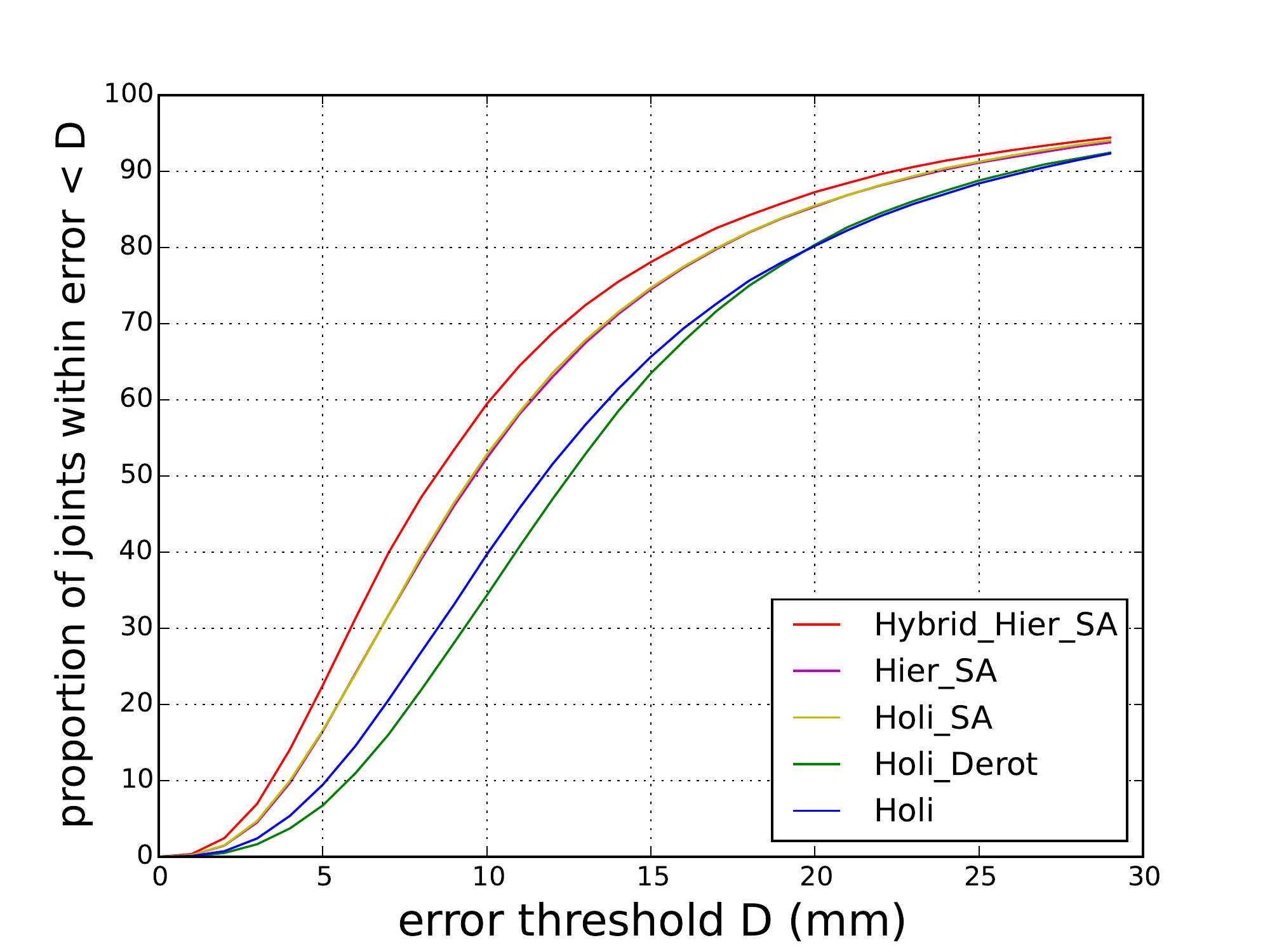}}
	\end{minipage}
	\begin{minipage}[b]{.325\linewidth}
	{\includegraphics[width=4.cm,trim=0.5cm 0cm 1.8cm 1.3cm,clip=true]{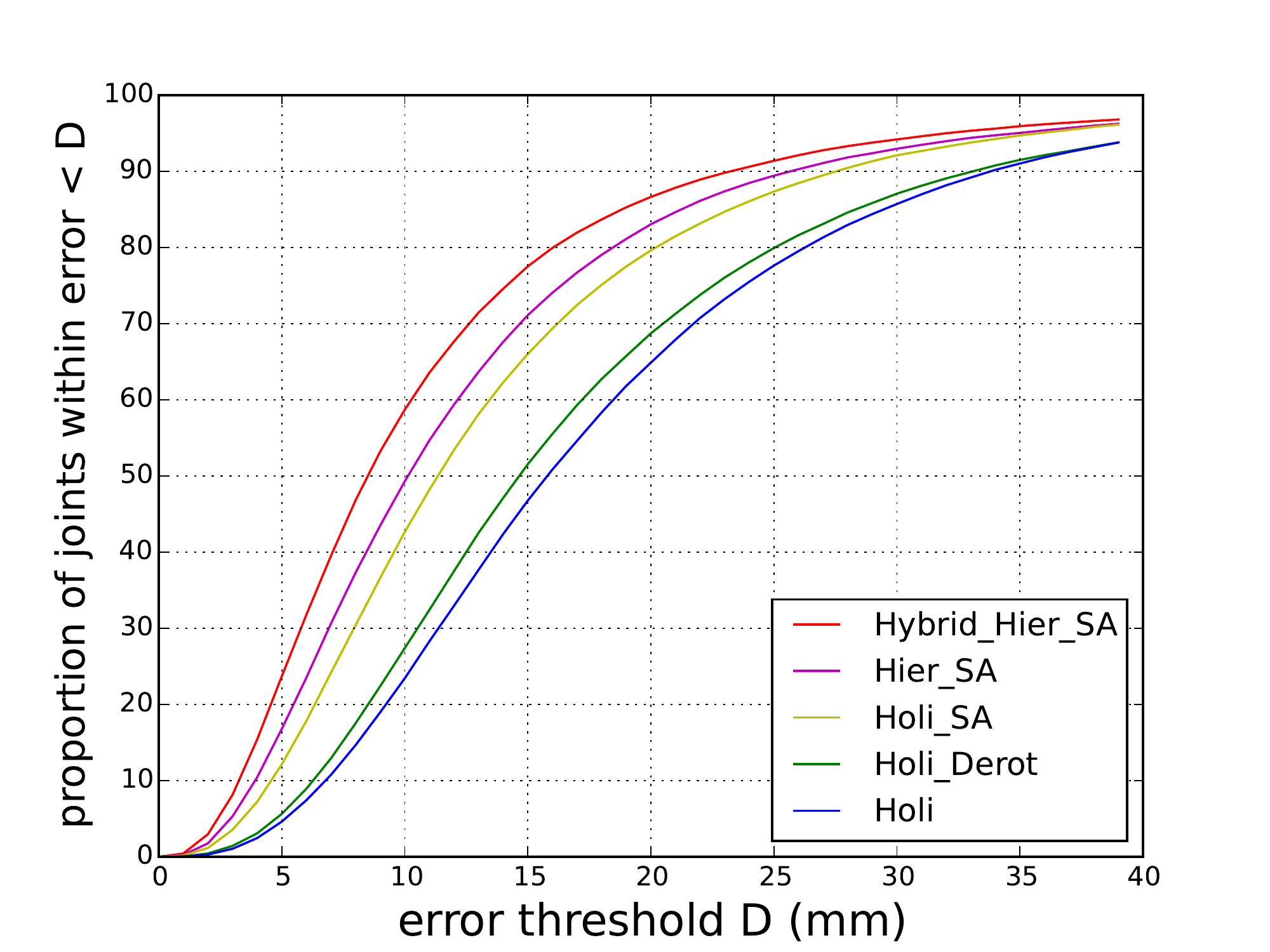}}
	\end{minipage}
	\begin{minipage}[b]{.325\linewidth}
	{\includegraphics[width=4.cm,trim=0.5cm 0cm 1.8cm 1.3cm,clip=true]{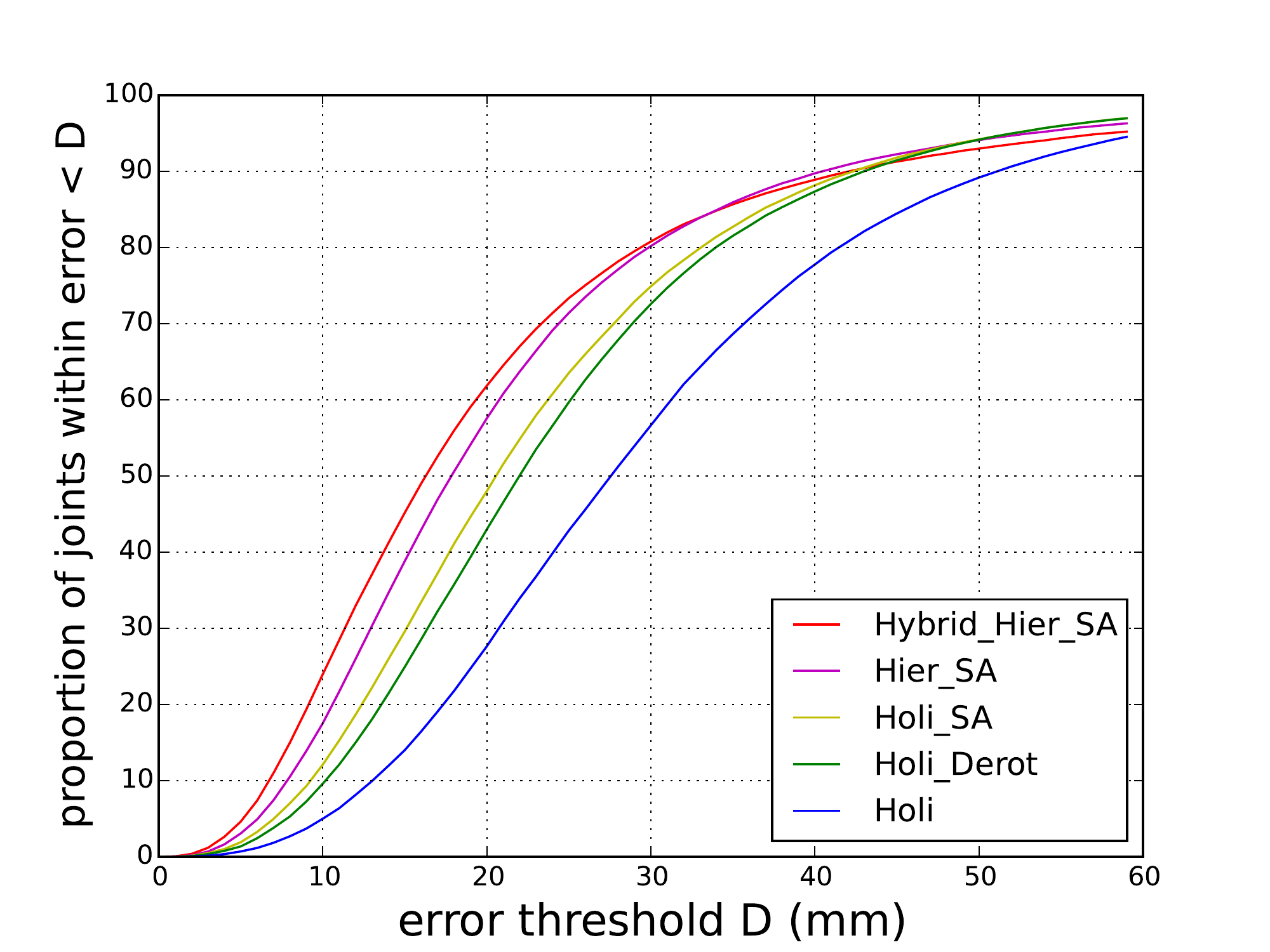}}
	\end{minipage}

\caption{Comparison of different methods on three datasets. Left:ICVL; Middle: NYU; Right: MSRC}
\label{fig:self}
\end{figure*}
On all the datasets, Hier_SA outperforms the baselines significantly, see Fig.~\ref{fig:self}. The improvement margin is related to the range of viewpoints and the complexity of articulations. The in-plane rotation distributions of original data, the ones after the  initial stage and the first stage are shown in Fig.~\ref{fig:iteration}. As MSRC dataset covers a full range of viewpoints and articulations, the improvement on this dataset from the baseline Holi is the largest.  For example, the percentages of frames under 20mm on ICVL, NYU and MSRC are improved by Hier_SA with margins of 5\%, 18\% and 30\% respectively, compared to that of Holi. 

The curves of Hier_SA and Holi_SA on three datasets illustrate the efficacy of hierarchical strategy in conquering the articulations, while the curves of Holi_SA, Holi and Holi_Derot show that spatial attention mechanism is effect in reducing the viewpoint and articulation complexity. By refining viewpoints with stages and spatially transforming the feature space to focus on the most relevant area for a certain joint estimation, Holi_SA achieves better results than Holi and  Holi_Derot. Note that the curve of Holi_Derot is under that of Holi on ICVL dataset, which implies that the error of estimating the rotation by a separate network may deteriorate the later estimation when the variations of the viewpoint in training set is small. 
 
Hybr_Hier_SA further improves the result of Hier_SA by a large margin consistently on all the datasets, which verifies that the kinematic constraints by the partial PSO is effective.

\subsection{Comparison with Prior Works}
\label{ssec:priorcmp}
\begin{figure*}
	\begin{minipage}[b]{.325\linewidth}
	{\includegraphics[width=4.cm,trim=0.5cm 0cm 1.8cm 1.3cm,clip=true]{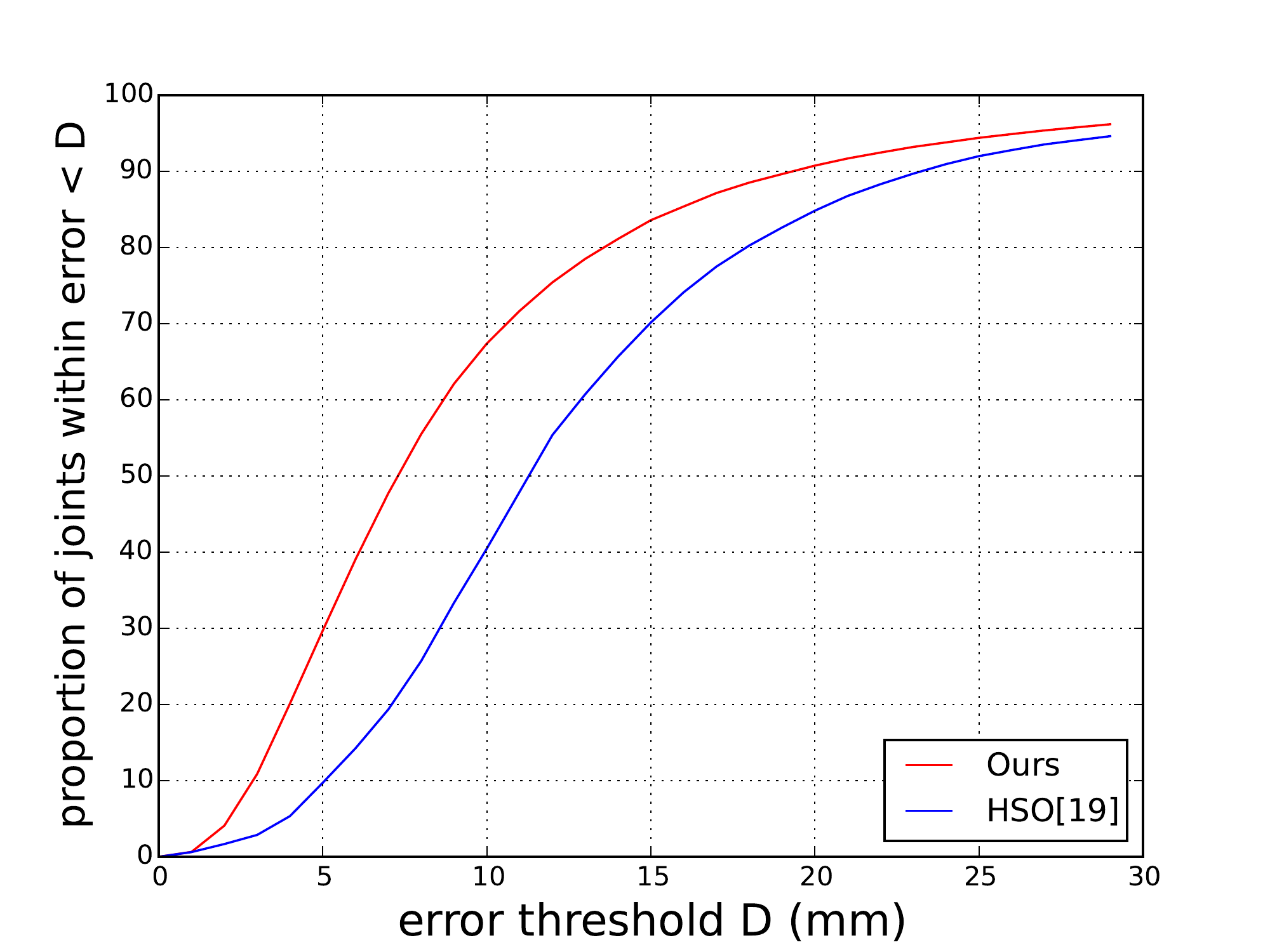}}
	\end{minipage}
	\begin{minipage}[b]{.325\linewidth}
	{\includegraphics[width=4.cm,trim=0.5cm 0cm 1.8cm 1.3cm,clip=true]{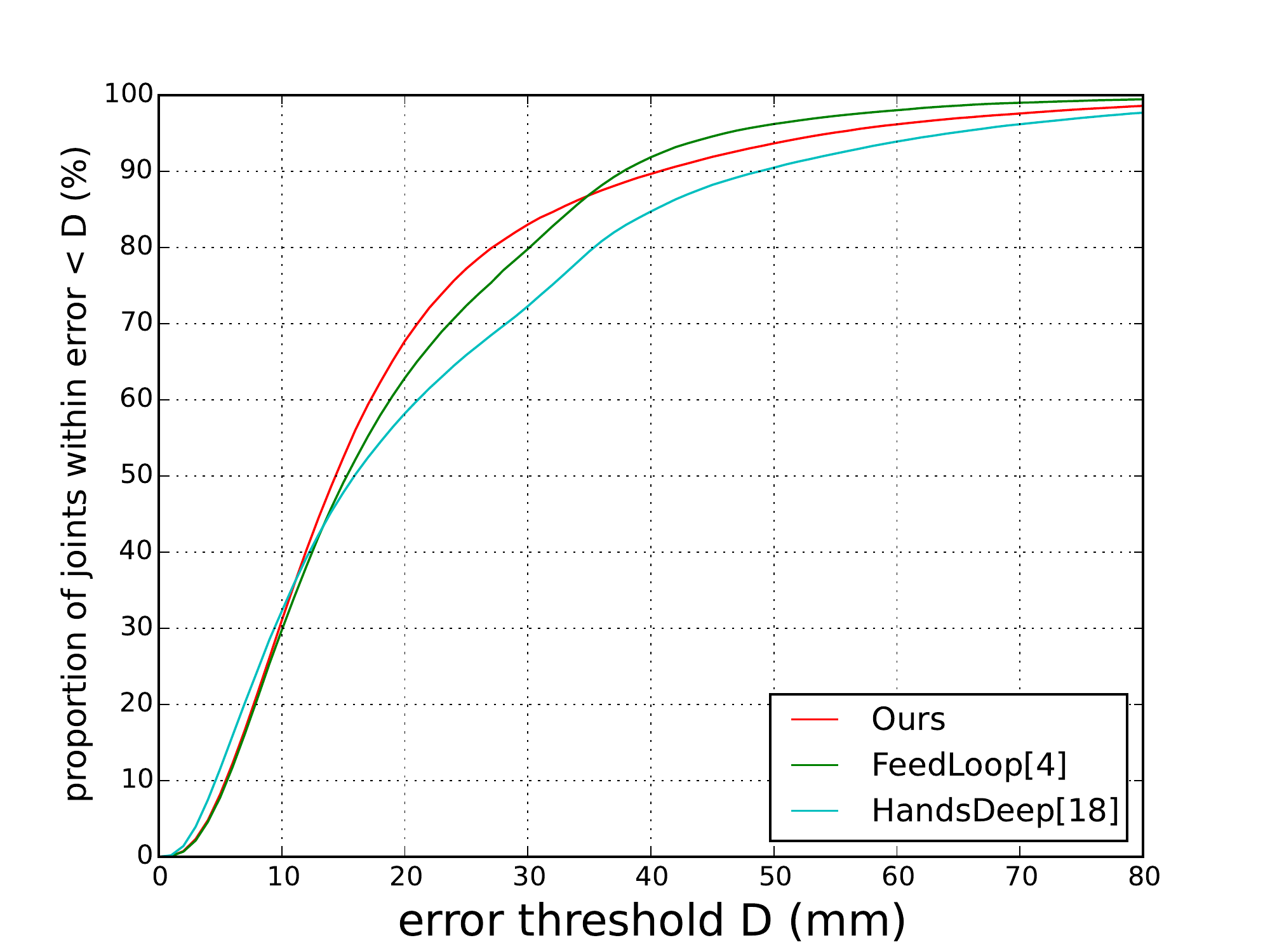}}
	\end{minipage}
	\begin{minipage}[b]{.325\linewidth}
	{\includegraphics[width=4cm,trim=0.5cm 0cm 1.8cm 1.3cm,clip=true]{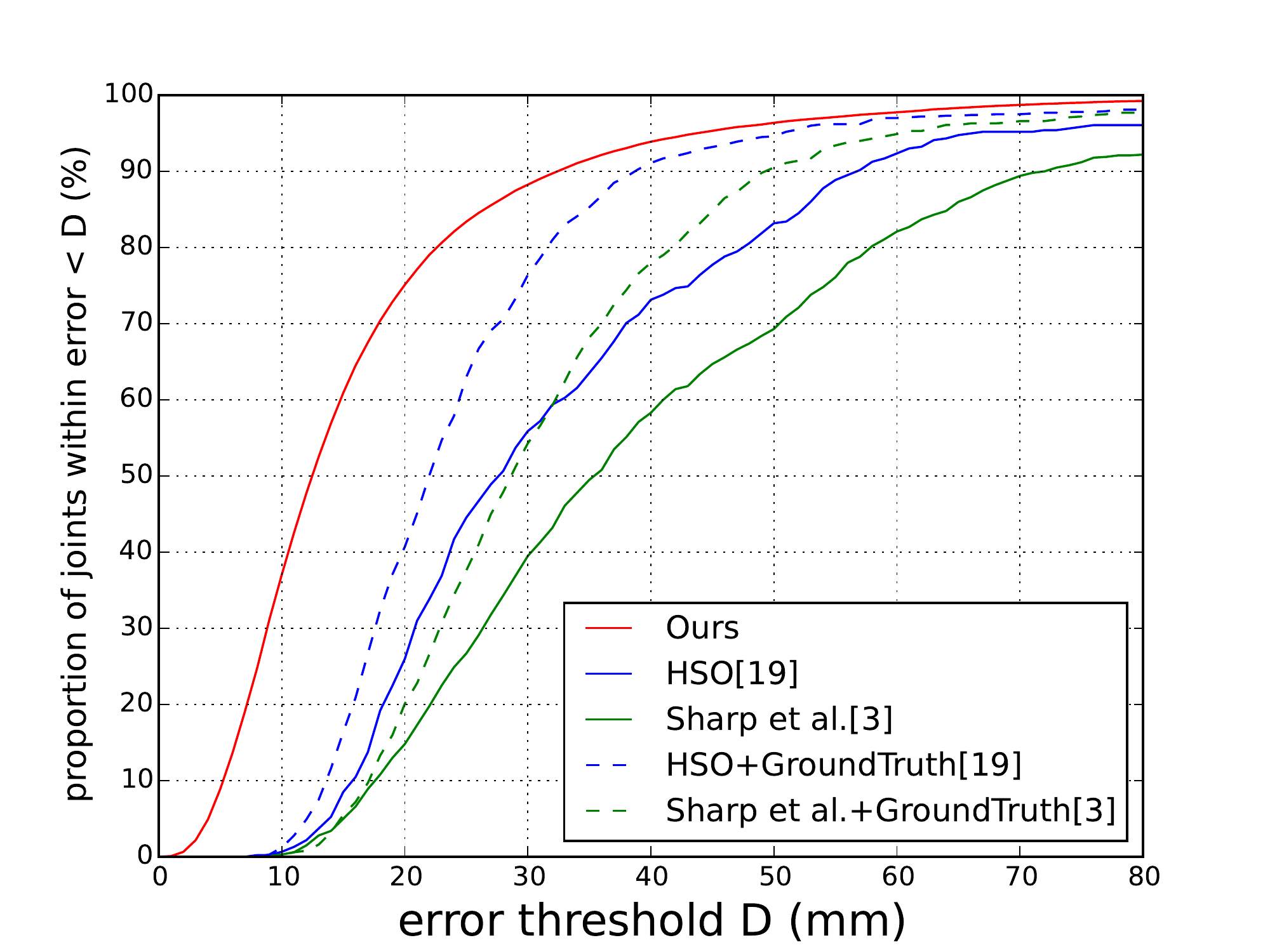}}
	\end{minipage}
\caption{Comparison of prior work on three datasets. Left:ICVL; Middle: NYU; Right: MSRC}
\label{fig:prior}
\end{figure*}

\begin{figure}[h]
	\centering
	
	\includegraphics[page=1,trim=0cm 0.5cm 0cm 0cm, clip = true, width=0.1\textwidth]{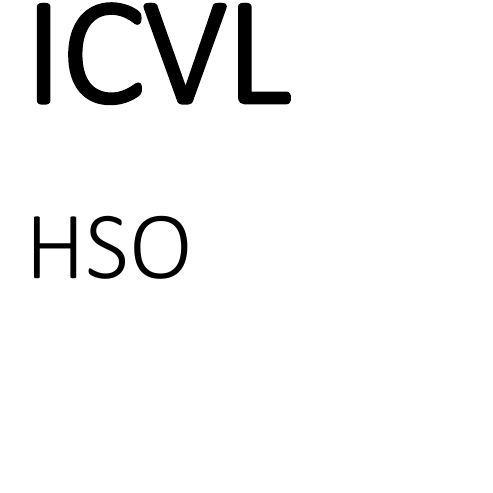}	
	\includegraphics[trim=16cm 13cm 16cm 10cm, clip = true, width=0.14\textwidth]{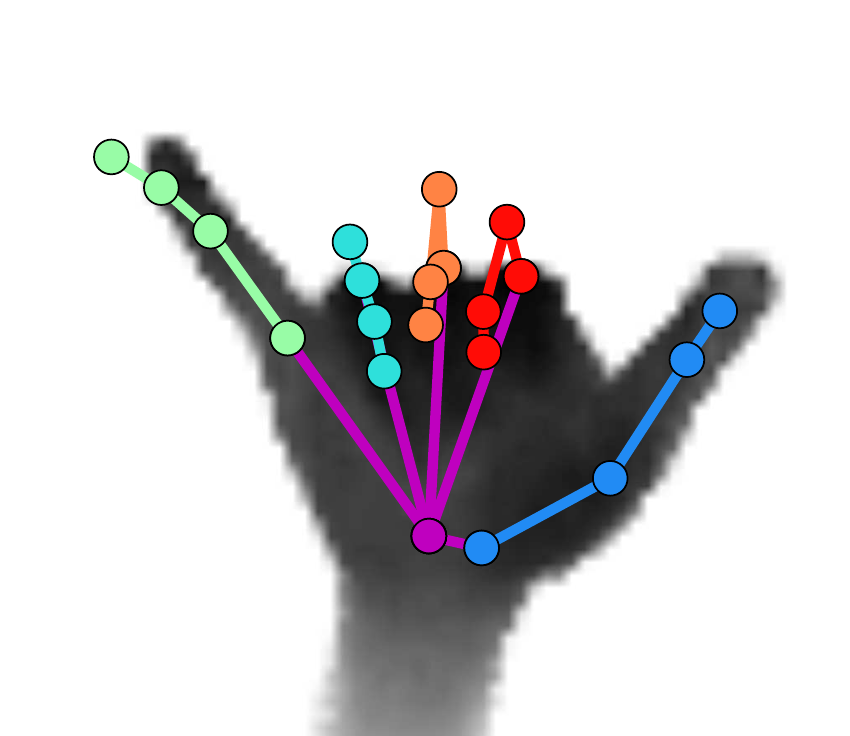}
	\includegraphics[trim=16cm 13cm 16cm 10cm, clip = true, width=0.14\textwidth]{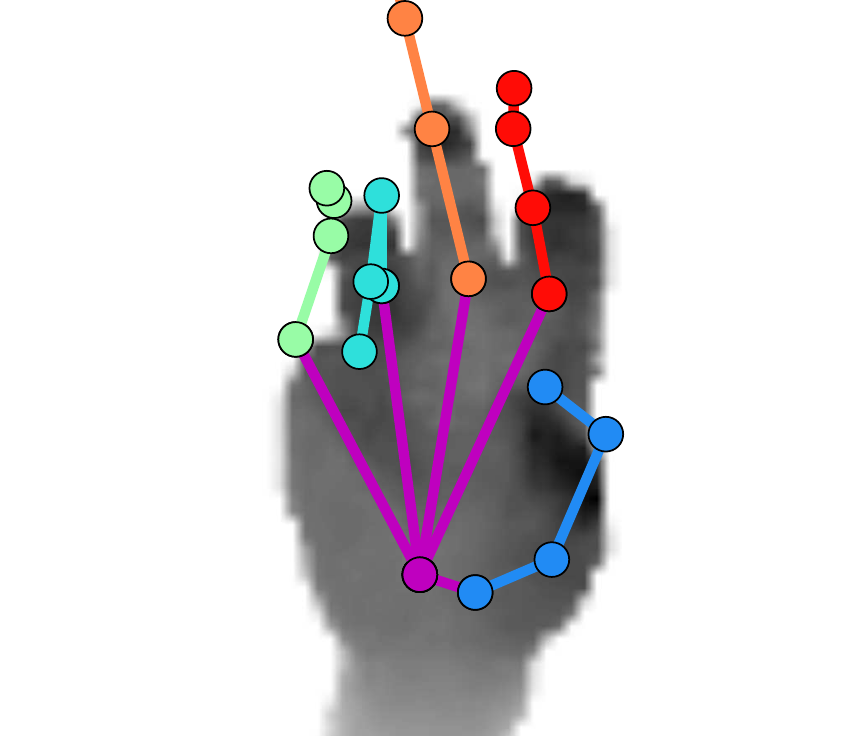}
	\includegraphics[trim=16cm 13cm 16cm 10cm, clip = true, width=0.14\textwidth]{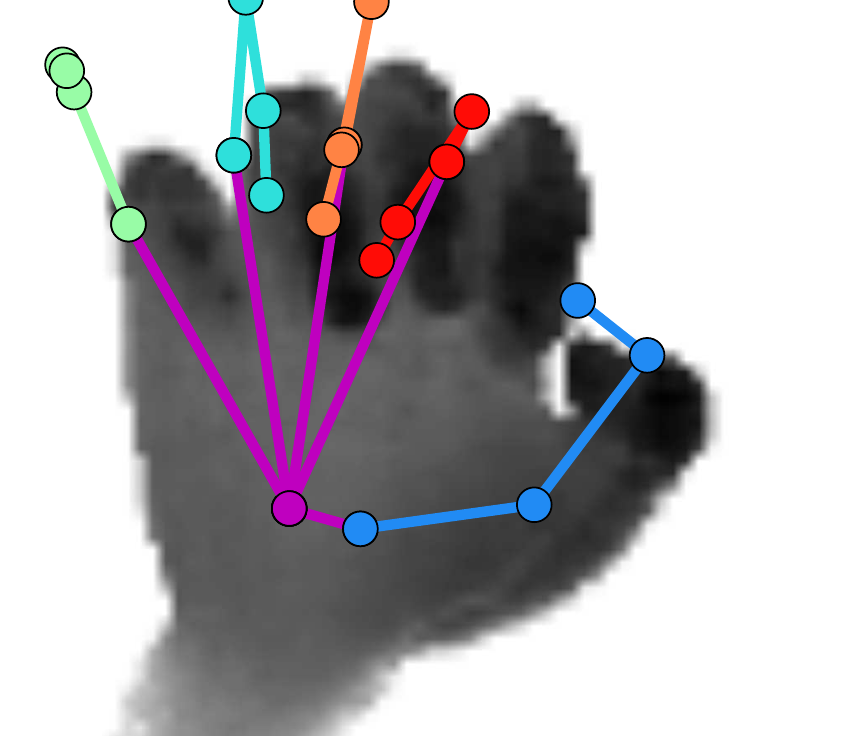}
	\includegraphics[trim=16cm 13cm 16cm 10cm, clip = true, width=0.14\textwidth]{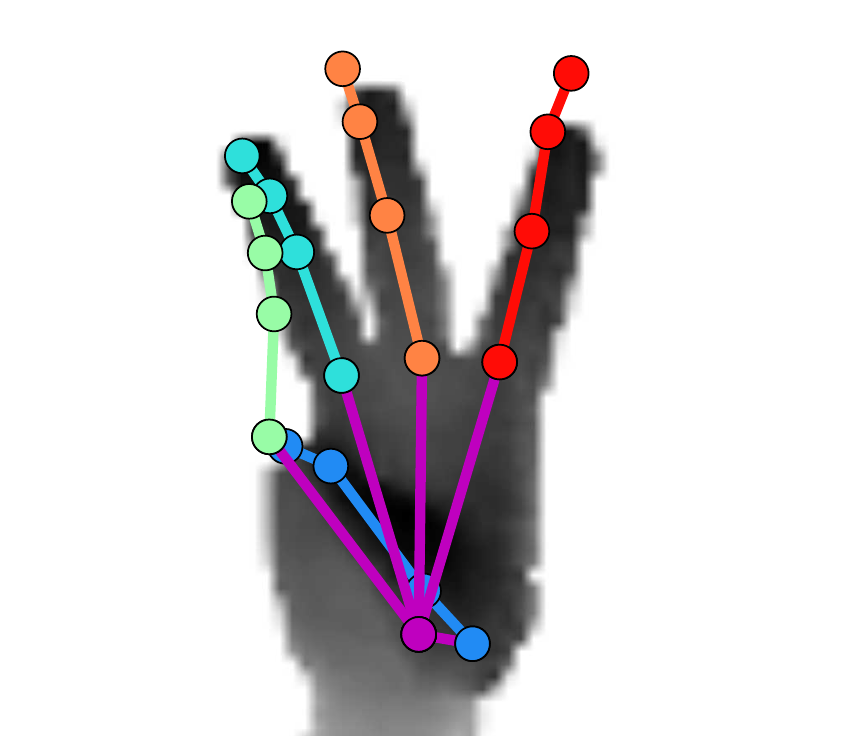}
	\includegraphics[trim=16cm 13cm 16cm 10cm, clip = true, width=0.14\textwidth]{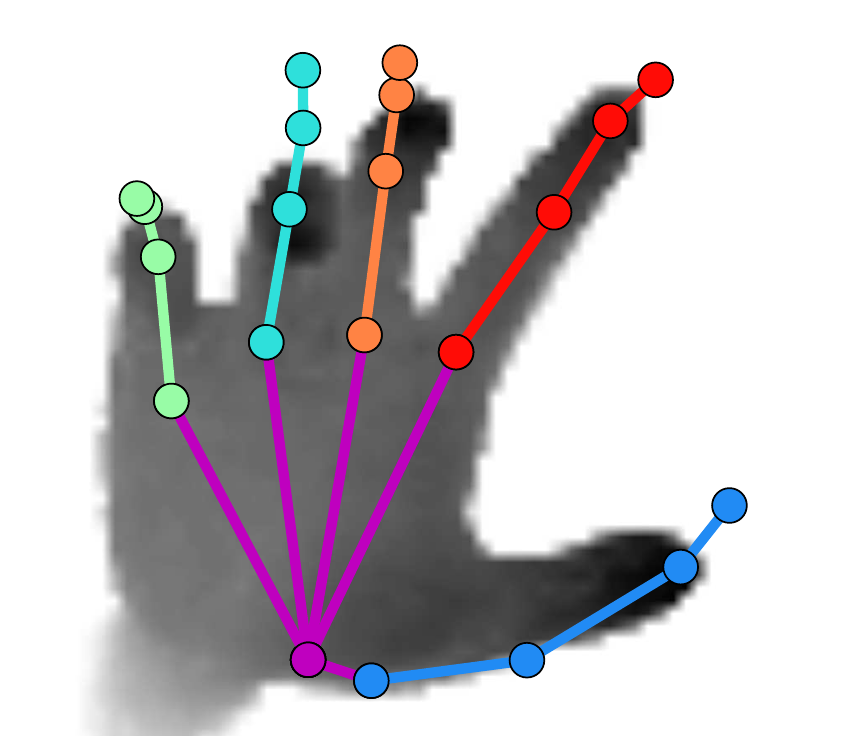}
	\includegraphics[trim=16cm 13cm 16cm 10cm, clip = true, width=0.14\textwidth]{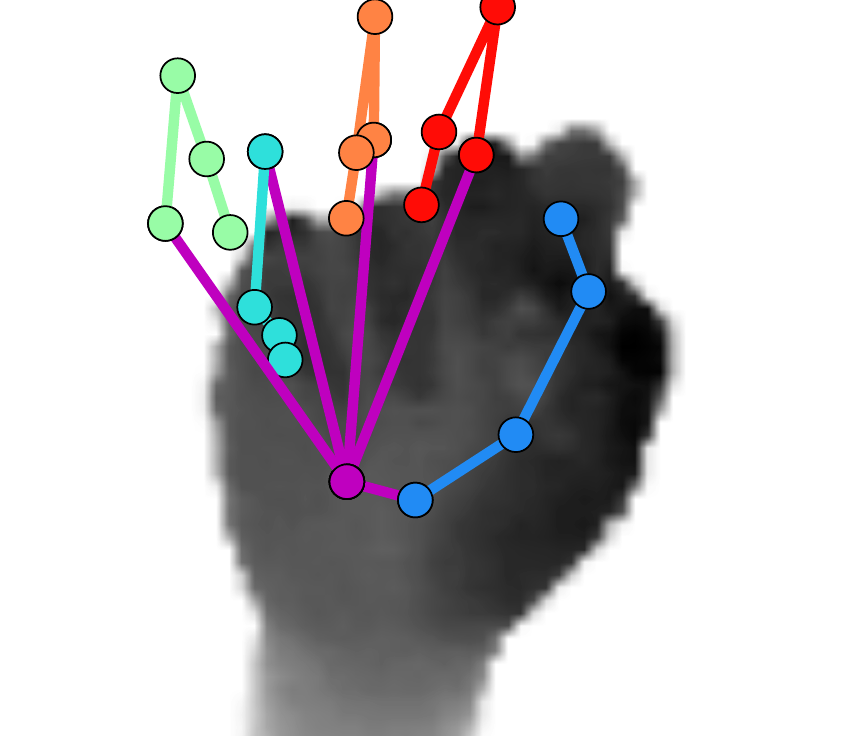}
	
	\includegraphics[page=5,trim=0cm 0.5cm 0cm 0cm, clip = true, width=0.1\textwidth]{text.pdf}						
	\includegraphics[trim=28cm 13cm 4cm 10cm, clip = true, width=0.14\textwidth]{selficvl0001-eps-converted-to.pdf}
	\includegraphics[trim=28cm 13cm 4cm 10cm, clip = true, width=0.14\textwidth]{selficvl0002-eps-converted-to.pdf}
	\includegraphics[trim=28cm 13cm 4cm 10cm, clip = true, width=0.14\textwidth]{selficvl0055-eps-converted-to.pdf}
	\includegraphics[trim=28cm 13cm 4cm 10cm, clip = true, width=0.14\textwidth]{selficvl0006-eps-converted-to.pdf}
	\includegraphics[trim=28cm 13cm 4cm 10cm, clip = true, width=0.14\textwidth]{selficvl0069-eps-converted-to.pdf}
	\includegraphics[trim=28cm 13cm 4cm 10cm, clip = true, width=0.14\textwidth]{selficvl0012-eps-converted-to.pdf}\\

	\includegraphics[page=2,trim=0cm 0.5cm 0cm 0cm, clip = true, width=0.1\textwidth]{text.pdf}	
	\includegraphics[trim=9.5cm 13cm 24cm 10cm, clip = true, width=0.14\textwidth]{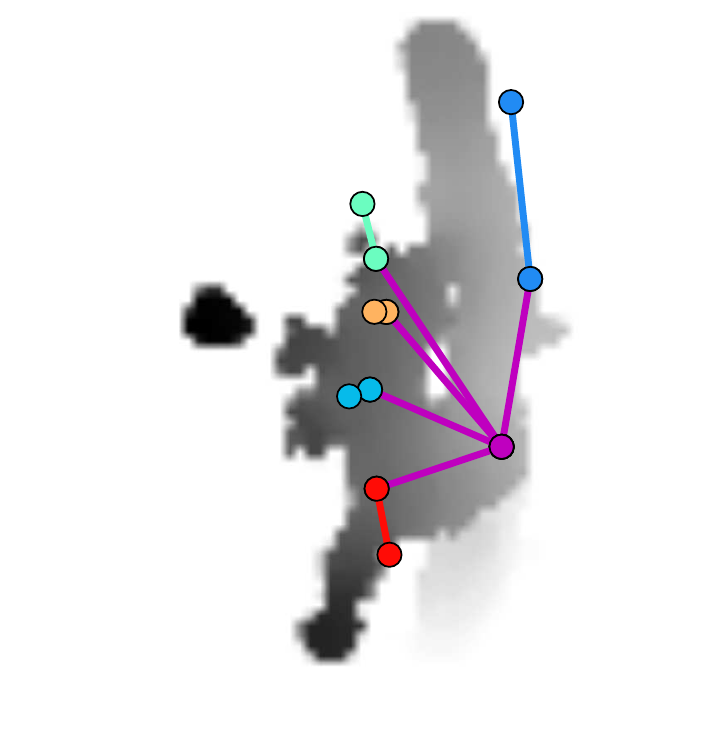}
	\includegraphics[trim=9.5cm 13cm 24cm 10cm, clip = true, width=0.14\textwidth]{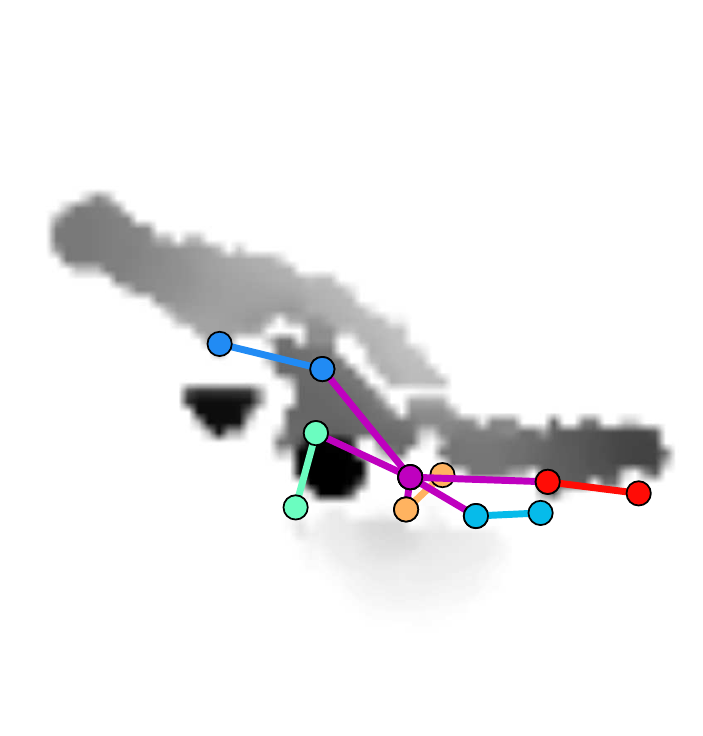}
	\includegraphics[trim=9.5cm 13cm 24cm 10cm, clip = true, width=0.14\textwidth]{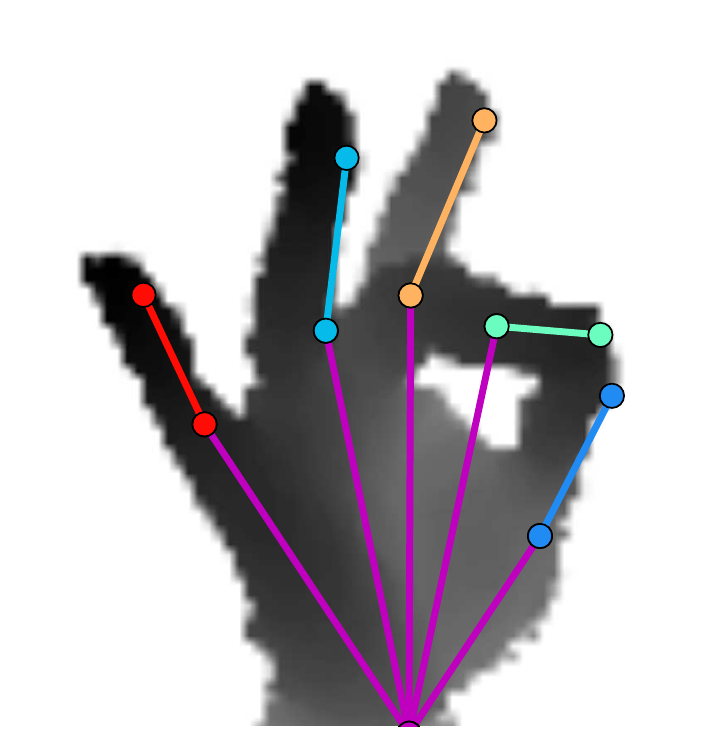}
	\includegraphics[trim=9.5cm 13cm 24cm 10cm, clip = true, width=0.14\textwidth]{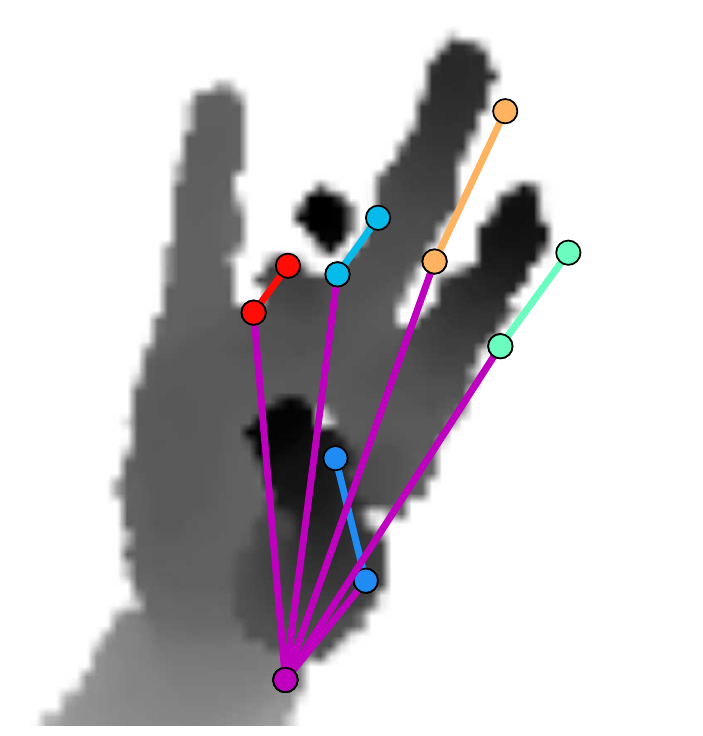}
	\includegraphics[trim=9.5cm 13cm 24cm 10cm, clip = true, width=0.14\textwidth]{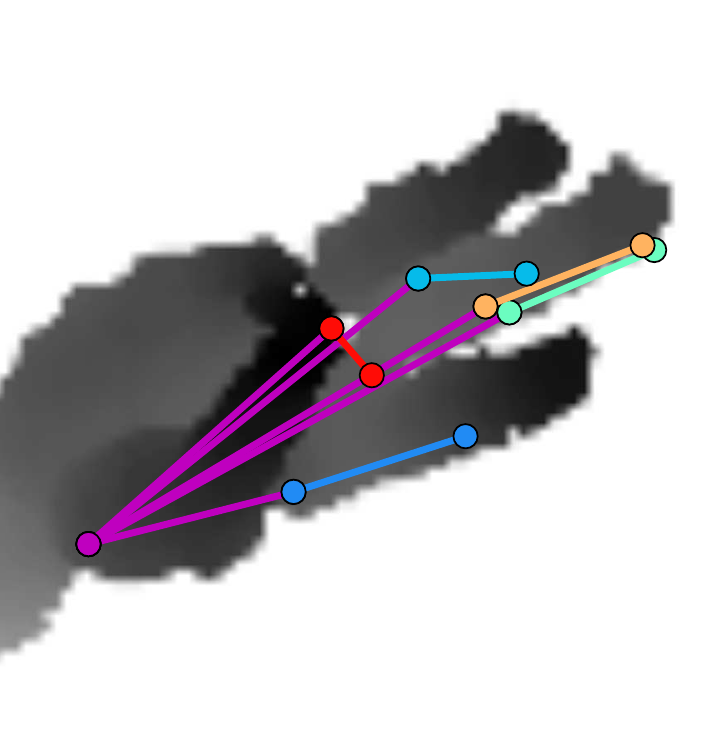}
	\includegraphics[trim=9.5cm 13cm 24cm 10cm, clip = true, width=0.14\textwidth]{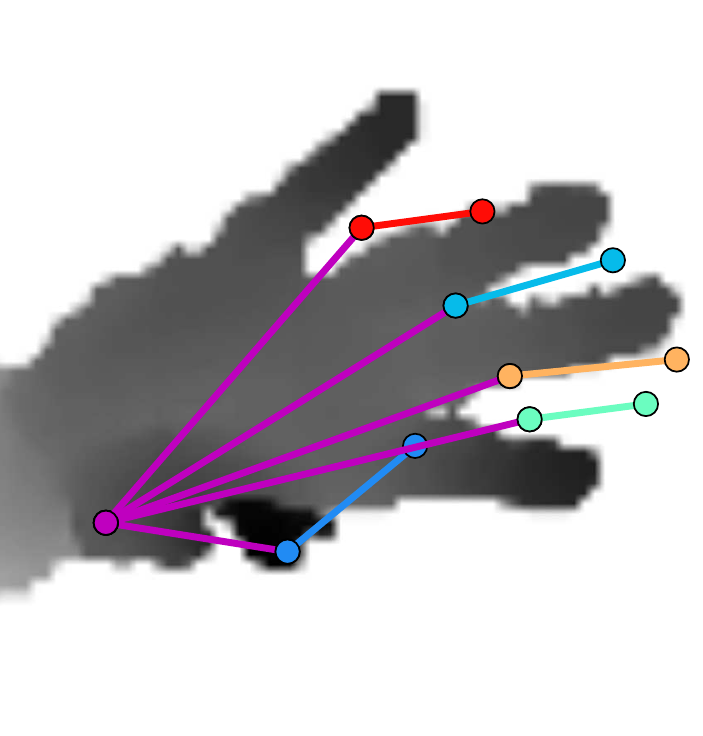}
	
	\includegraphics[page=3,trim=0cm 0.5cm 0cm 0cm, clip = true, width=0.1\textwidth]{text.pdf}						
	\includegraphics[trim=16.5cm 13cm  16.8cm 10cm, clip = true, width=0.14\textwidth]{selfnyu0024-eps-converted-to.pdf}
	\includegraphics[trim=16.5cm 13cm 16.8cm 10cm, clip = true, width=0.14\textwidth]{selfnyu0026-eps-converted-to.pdf}
	\includegraphics[trim=16.5cm 13cm 16.8cm 10cm, clip = true, width=0.14\textwidth]{selfnyu0039-eps-converted-to.pdf}
	\includegraphics[trim=16.5cm 13cm 16.8cm 10cm, clip = true, width=0.14\textwidth]{selfnyu0044-eps-converted-to.pdf}
	\includegraphics[trim=16.5cm 13cm 16.8cm 10cm, clip = true, width=0.14\textwidth]{selfnyu0049-eps-converted-to.pdf}
	\includegraphics[trim=16.5cm 13cm 16.8cm 10cm, clip = true, width=0.14\textwidth]{selfnyu0050-eps-converted-to.pdf}\\
	
	\includegraphics[page=5,trim=0cm 0.5cm 0cm 0cm, clip = true, width=0.1\textwidth]{text.pdf}						
	\includegraphics[trim=24.5cm 13cm 9.5cm 10cm, clip = true, width=0.14\textwidth]{selfnyu0024-eps-converted-to.pdf}
	\includegraphics[trim=24.5cm 13cm 9.5cm 10cm, clip = true, width=0.14\textwidth]{selfnyu0026-eps-converted-to.pdf}
	\includegraphics[trim=24.5cm 13cm 9.5cm 10cm, clip = true, width=0.14\textwidth]{selfnyu0039-eps-converted-to.pdf}
	\includegraphics[trim=24.5cm 13cm 9.5cm 10cm, clip = true, width=0.14\textwidth]{selfnyu0044-eps-converted-to.pdf}
	\includegraphics[trim=24.5cm 13cm 9.5cm 10cm, clip = true, width=0.14\textwidth]{selfnyu0049-eps-converted-to.pdf}
	\includegraphics[trim=24.5cm 13cm 9.5cm 10cm, clip = true, width=0.14\textwidth]{selfnyu0050-eps-converted-to.pdf}\\	
	
	\includegraphics[page=4,trim=0cm 0.5cm 0cm 0cm, clip = true, width=0.1\textwidth]{text.pdf}	
	\includegraphics[trim=16cm 13cm 16cm 10cm, clip = true, width=0.14\textwidth]{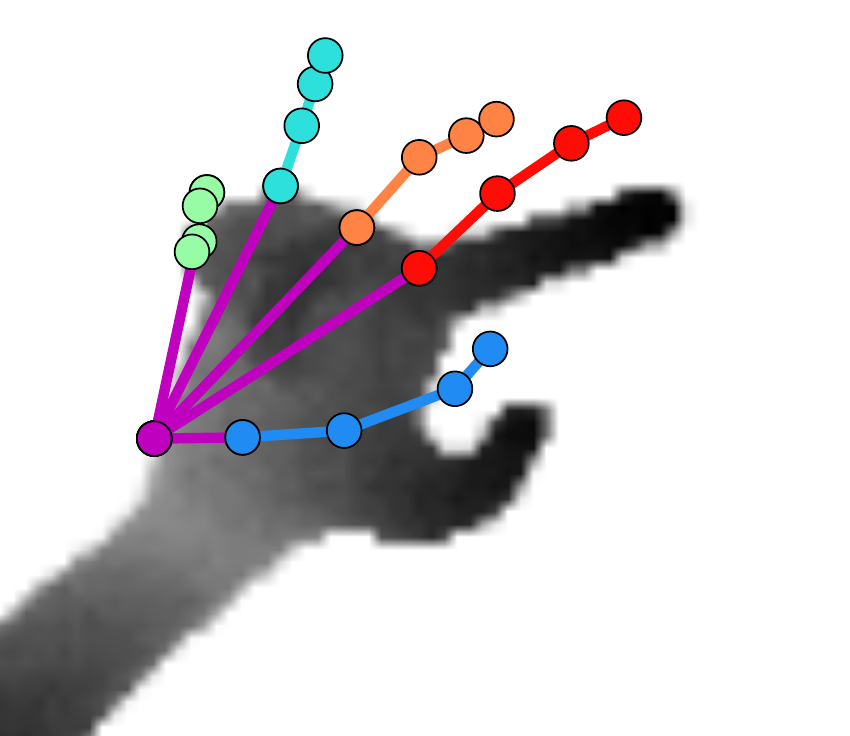}
	\includegraphics[trim=16cm 13cm 16cm 10cm, clip = true, width=0.14\textwidth]{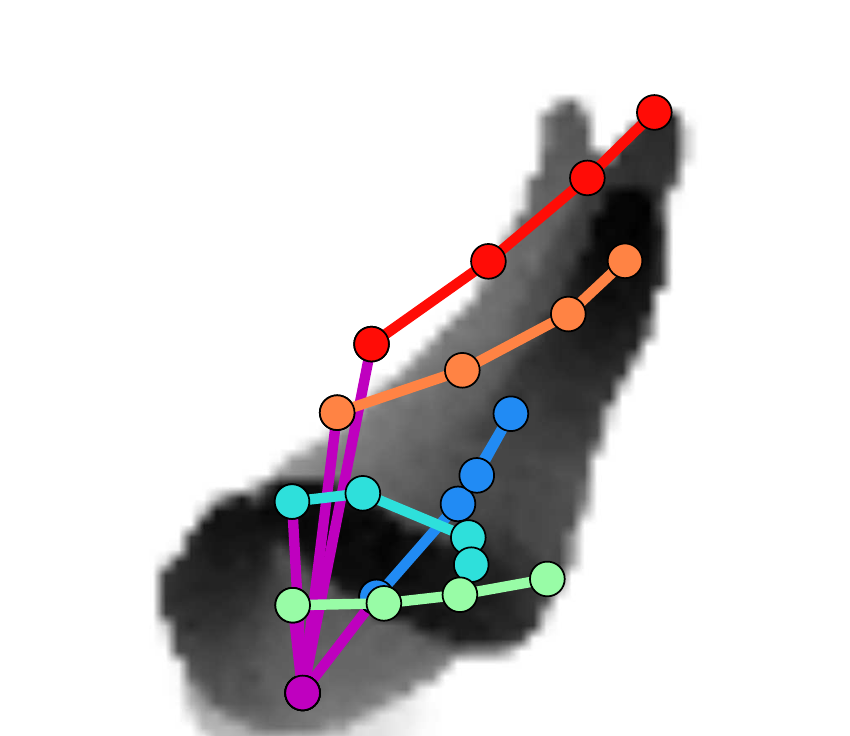}
	\includegraphics[trim=16cm 13cm 16cm 10cm, clip = true, width=0.14\textwidth]{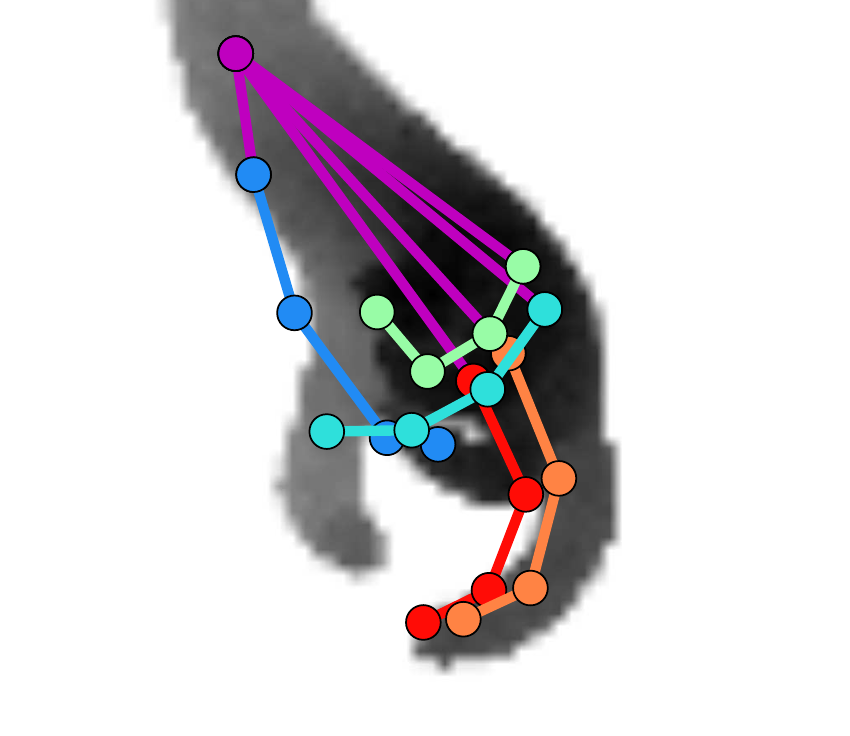}
	\includegraphics[trim=16cm 13cm 16cm 10cm, clip = true, width=0.14\textwidth]{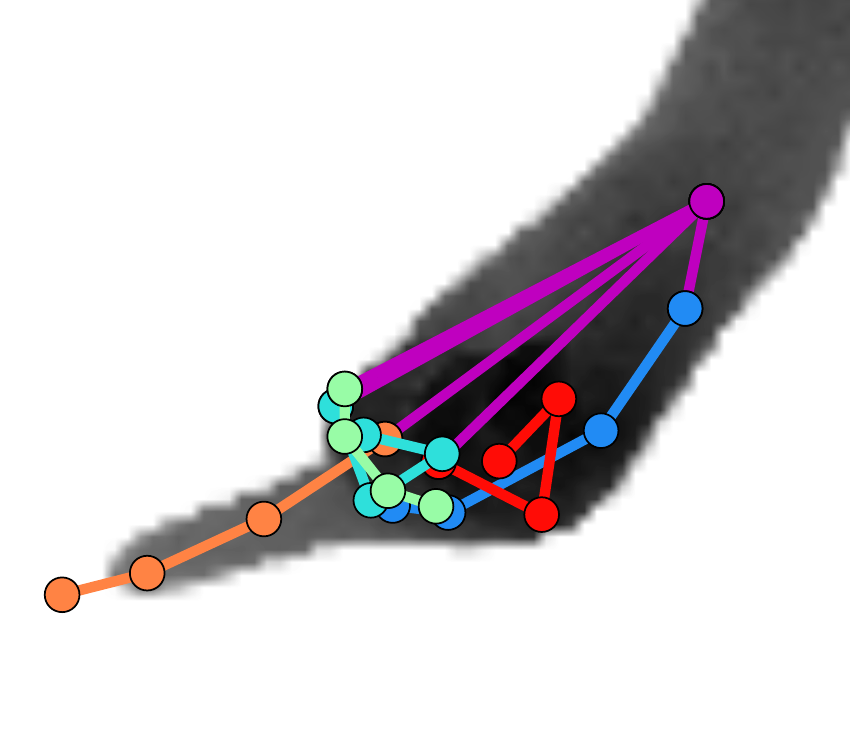}
	\includegraphics[trim=16cm 13cm 16cm 10cm, clip = true, width=0.14\textwidth]{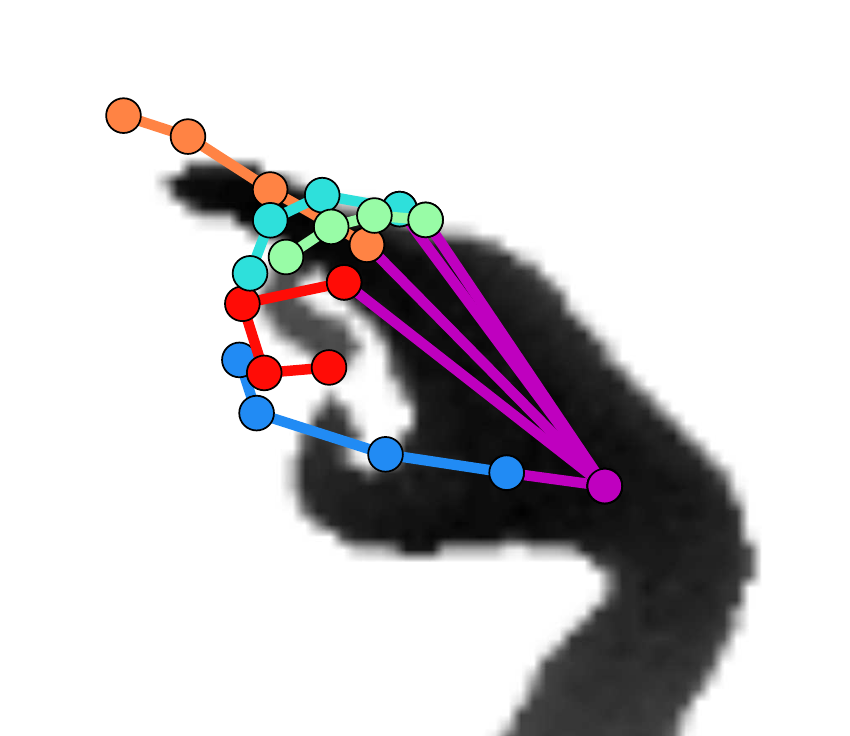}
	\includegraphics[trim=16cm 13cm 16cm 10cm, clip = true, width=0.14\textwidth]{selficvl0012-eps-converted-to.pdf}

	\includegraphics[page=5,trim=0cm 0.5cm 0cm 0cm, clip = true, width=0.1\textwidth]{text.pdf}						
	\includegraphics[trim=28cm 13cm 4cm 10cm, clip = true, width=0.14\textwidth]{selfmsrc0003-eps-converted-to.pdf}
	\includegraphics[trim=28cm 13cm 4cm 10cm, clip = true, width=0.14\textwidth]{selfmsrc0004-eps-converted-to.pdf}
	\includegraphics[trim=28cm 13cm 4cm 10cm, clip = true, width=0.14\textwidth]{selfmsrc0006-eps-converted-to.pdf}
	\includegraphics[trim=28cm 13cm 4cm 10cm, clip = true, width=0.14\textwidth]{selfmsrc0007-eps-converted-to.pdf}
	\includegraphics[trim=28cm 13cm 4cm 10cm, clip = true, width=0.14\textwidth]{selfmsrc0010-eps-converted-to.pdf}
	\includegraphics[trim=28cm 13cm 4cm 10cm, clip = true, width=0.14\textwidth]{selficvl0012-eps-converted-to.pdf}\\
		
	\caption{Examples comparing to prior work on three datasets. The first two rows, the middle  three rows, and the last two rows are examples from ICVL dataset, NYU dataset and MSRC dataset, respectively.  Compared to other methods, our method has a good performance in discriminating the fingers and a better precision. For many challenging viewpoints, our method still has a good estimation.}
\label{fig:quacmpprior}
\end{figure}

We compare our work with 4 state-of-the-arts methods: Hierarchical Sampling Optimization(HSO)~\cite{tang2015opening}, Sharp {\it et al.}~\cite{sharp2015accurate}, HandsDeep~\cite{oberweger2015hands}, FeedLoop~\cite{oberweger2015training} on three datasets, see Fig.~\ref{fig:prior}.  The former two are hybrid methods and the the latter two are refinement method based on CNN. The results are obtained either from the authors for HSO~\cite{tang2015opening} or from the reported accuracies ~\cite{sharp2015accurate,oberweger2015hands,oberweger2015training}. The examples of the estimation results of HandsDeep, FeedLoop, HSO and our method are shown in Fig.~\ref{fig:quacmpprior}.

On ICVL dataset, we compare HSO with parameters set as $N=100,M=150$. Our method is better by 26\% of joints  within $D=10mm$. On NYU dataset, we compare our method with HandsDeep and FeedLoop which are all based on CNN. As the hand model of these methods are different, we evaluate the result by comparing the error of the subset of 11 joint locations(removing the palm joints except the root joint of thumb). Our estimation result is better than HandsDeep by a large margin, for example, an improvement of 10 \% within  $D=30mm$, and achieves roughly the same accuracy with FeedLoop.

We finally test our method with HSO and Sharp {\it et al.} on MSRC dataset. The dataset is more challenging than the above two as it covers a wider range of viewpoints and articulations. The curves demonstrate the superiority of our method under large variations. For example, the proportion of joints (when $D=30mm$) of our method is 35\% and 50\% more than those of HSO and Sharp {\it et al.} respectively. Note that our estimation is even better than the results of HSO and Sharp {\it et al.} using ground truth rotation~\cite{tang2015opening}.

\section{Conclusion}
\label{sec:conclusion}
To apply the hierarchy strategy to the input and feature space and enforce the hand kinematic constraints to the hand pose estimation, we present a hybrid method by applying the kinematic hierarchy to both the input and feature space of the discriminative method and the optimization of the generative method. For the integration of hierarchical input and feature space of the discriminative, a spatial attention mechanism is introduced to spatially transform the input(and feature) and output space interactively, leading to new spaces with lesser viewpoint and articulation complexity and gradually refining th estimation results. In addition, the partial PSO is incorporated between the layers of the hierarchy to enforce the kinematic constraints to the estimation results of the discriminative part. This helps reduce the error from previous layer to accumulate. Our method demonstrates good performance on three datasets, especially on the dataset under large variations.

\clearpage
\bibliographystyle{splncs}
\bibliography{egbib}

\end{document}